\documentclass[5p,times]{elsarticle}
\usepackage{amssymb}
\usepackage{amsmath}
\usepackage{lineno}
\usepackage{multirow}
\usepackage{tabularx}
\usepackage{makecell}
\usepackage{float}
\usepackage{wrapfig}
\usepackage[caption=false,font=footnotesize]{subfig}
\usepackage{graphicx}
\usepackage{adjustbox}
\usepackage{xcolor}
\usepackage{soul}
\usepackage{siunitx}
\usepackage{booktabs}
\journal{Mechatronics}
\usepackage{graphicx}

\begin{document}
\begin{frontmatter}

\title{Learning-Based Modeling of a Magnetically Steerable Soft Suction Device for Endoscopic Endonasal Interventions}

\author[aff1]{Majid Roshanfar\corref{cor1}}
\ead{majid.roshanfar@sickkids.ca}

\author[aff2]{Alex Zhang}

\author[aff3]{Changyan He}

\author[aff4]{Amir Hooshiar}

\author[aff1]{Dale J. Podolsky}

\author[aff1]{Thomas Looi}

\author[aff2,aff5]{Eric Diller}

\cortext[cor1]{Corresponding author.}

\affiliation[aff1]{organization={The Wilfred and Joyce Posluns Centre for Image Guided Innovation \& Therapeutic Intervention (PCIGITI), The Hospital for Sick Children (SickKids)},
  city={Toronto},
 state={ON},
  country={Canada}}

\affiliation[aff2]{organization={Department of Mechanical and Industrial Engineering (MIE), University of Toronto},
  city={Toronto},
  state={ON},
  country={Canada}}

\affiliation[aff3]{organization={School of Engineering, University of Newcastle},
  city={Newcastle},
  state={NSW},
  country={Australia}}

\affiliation[aff4]{organization={Surgical Performance Enhancement and Robotics (SuPER) Centre, Department of Surgery, McGill University},
  city={Montreal},
  state={QC},
  country={Canada}}

\affiliation[aff5]{organization={Institute of Biomedical Engineering (BME), University of Toronto},
  city={Toronto},
  state={ON},
  country={Canada}}

\begin{abstract}
This paper introduces a learning-based modeling framework for a magnetically steerable soft suction device designed for endoscopic endonasal brain tumor resection. The device is miniaturized (4~mm outer diameter, 2~mm inner diameter, 40~mm length), 3D printed using biocompatible SIL 30 material, and integrates embedded Fiber Bragg Grating (FBG) sensors for real-time shape feedback. Shape reconstruction is represented using four Bezier control points, allowing for a compact and smooth representation of the device's deformation.
{A data-driven model was trained on 5,097 experimental samples to learn the mapping from magnetic field parameters (magnitude: 0--14~mT, frequency: 0.2--1.0~Hz, and vertical tip distances from the surface of the electromagnet coil table: 90--100~mm) to the resulting geometric configuration of the soft robot, represented by four Bezier control points that define its 3D shape. The model was implemented and compared using both Neural Network (NN) and Random Forest (RF) architectures.}
The RF model outperformed the NN across all metrics, achieving a mean root mean square error of 0.087~mm in control point prediction and a mean shape reconstruction error of 0.064~mm. Feature importance analysis further revealed that magnetic field components predominantly influence distal control points, while frequency and distance affect the base configuration. {Unlike prior studies that apply general machine learning methods to soft robotic data, the proposed framework introduces a new modeling paradigm that links magnetic actuation inputs directly to geometric Bezier control points, creating an interpretable and low-dimensional representation of deformation.} {This conceptual integration of magnetic field characterization, embedded FBG sensing, and Bezier-based learning provides a unified modeling strategy that can be extended to other magnetically actuated continuum robots.} This learning-based approach effectively models the complex nonlinear behavior of hyperelastic soft robots under magnetic actuation without relying on simplified physical assumptions. By enabling sub-millimeter shape prediction accuracy and real-time inference, this work establishes an advancement toward the intelligent control of magnetically actuated soft robotic tools in minimally invasive neurosurgery.
\end{abstract}



\end{frontmatter}

\section{Introduction}
\label{Introduction}
\begin{figure}[!t]
  \centering
    \includegraphics[width=\linewidth]{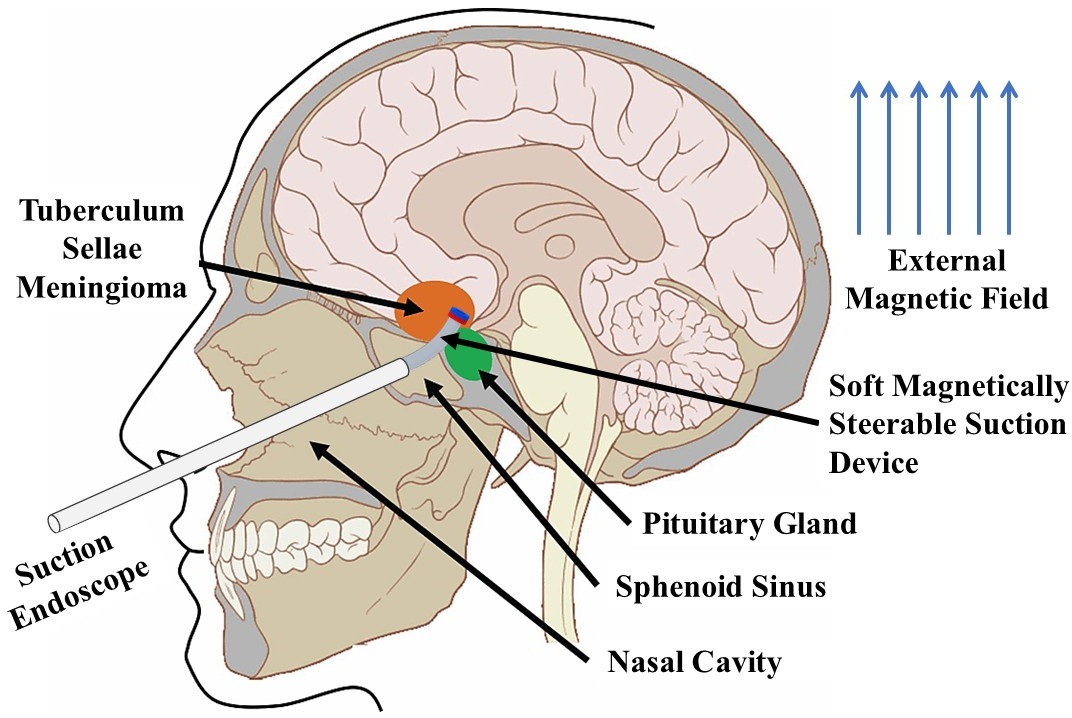}\vspace{5 mm}
    \includegraphics[width=\linewidth]{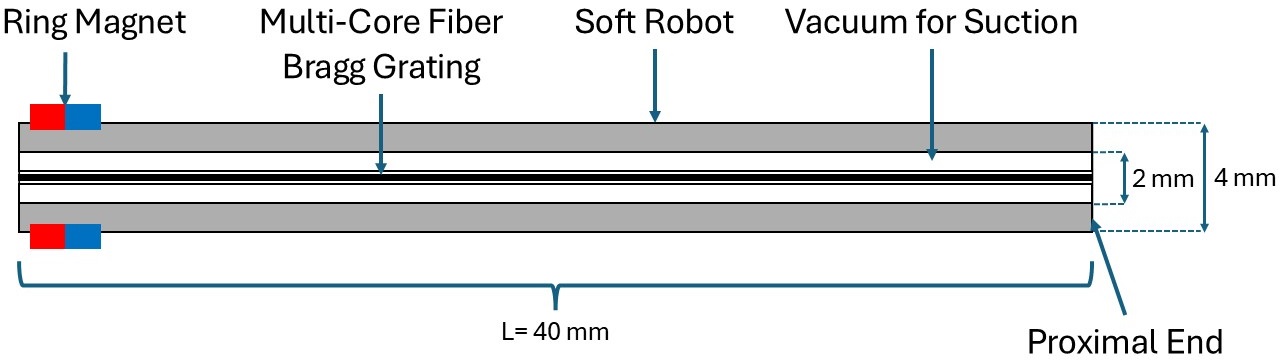}
    \caption{Magnetically steerable soft suction device for endoscopic endonasal brain tumor resection.
   (Top) Conceptual illustration showing how the device is inserted through the nasal cavity and sphenoid sinus using a standard suction endoscope to reach deep-seated skull-base regions. 
   {(Bottom) Cross-sectional schematic illustrating the device’s mechanical design and dimensions. The soft body (gray) encloses a suction lumen ({2}{mm}) and a channel routing the multi-core Fiber Bragg Grating (FBG) sensor at the neutral axis for accurate curvature sensing. The outer diameter is {4}{mm}, and the active flexible length is {40}{mm}. A hollow cylindrical permanent magnet (red/blue) at the distal tip enables magnetic steering under externally applied rotating magnetic fields. Additional geometric and material specifications are provided in Table}~\ref{tab:device_specs}.}
    \label{fig:01}
\end{figure}

An endoscopic endonasal approach is a well-established, minimally invasive alternative to traditional craniotomy, aimed at reducing trauma to the brain and surrounding critical neurovascular structures~\cite{martinez2021modern}. By making use of natural anatomical pathways through the nasal cavity, this technique eliminates the need for large scalp incisions or bone removal, reducing patient complications and recovery time~\cite{gholami2025advances}. It involves removing a portion of the sinus cavity using a surgical drill to create a narrow surgical corridor for the insertion of two primary instruments: an endoscopic camera for high-definition visualization and a surgical tool, typically a suction or cutting device, for tissue manipulation. Together, these instruments enable surgeons to operate within deep and confined spaces, performing precise dissection and controlled, stepwise removal of tumor tissue under continuous visual guidance (see Fig.~\ref{fig:01}). Clinical studies have demonstrated that this approach offers favorable neurological outcomes, shorter hospital stays, and reduced postoperative complications, particularly in pediatric patients with intracranial and skull base tumors~\cite{sasagawa2024endoscopic, kim2019endoscopic}.

However, the effectiveness of the endoscopic endonasal approach is often constrained by patient-specific anatomical complexities, such as a reduced skull base, narrow nasal passages, and underdeveloped sphenoid sinuses~\cite{valencia2024special}. These anatomical challenges significantly restrict the maneuverability of conventional rigid and straight instruments, increasing the risk of unintentional damage to critical structures like blood vessels, optic nerves, and the brainstem. As a result, surgical precision, safety, and access to certain tumor locations can be compromised. To address these limitations, flexible robotic instruments have been introduced~\cite{roshanfar2022stiffness}, leveraging the compliance and adaptability of their materials to navigate complex and curved anatomical pathways~\cite{burgner2013telerobotic, rox2020mechatronic}. These systems aim to minimize tissue trauma while enhancing dexterity, reach, and control within confined surgical environments. Nevertheless, the inherent nonlinear and elastic behavior of flexible robotic instruments, combined with the lack of intuitive control and real-time shape feedback, continues to pose significant challenges for safe and effective clinical use.

{Traditional analytical models and constant-curvature {kinematic} formulations have been widely used to {derive closed-form relationships between actuation inputs and the resulting configuration of} continuum and soft robots~\cite{webster2010design, rucker2011statics}. {The constant-curvature assumption, as systematically formalized by Webster and Jones}~\cite{webster2010design}{, provides computationally efficient closed-form solutions by treating each robot segment as a circular arc with uniform bending. Godage et al.}~\cite{godage2015modal} {extended this framework through modal kinematics for multisection continuum arms, enabling coordinated control of multiple bending segments.} While effective for simple bending scenarios, these models rely on simplifying assumptions such as uniform curvature, isotropic stiffness, and quasi-static actuation. {More advanced physics-based frameworks have been developed to address these limitations. Cosserat rod theory, as applied to continuum robots by Rucker and Webster}~\cite{rucker2011statics}{, provides geometrically exact formulations that capture all six deformation modes (bending about two axes, torsion, extension, and shear about two axes). Till et al.}~\cite{till2019real} {demonstrated real-time dynamics simulation of Cosserat-based models using implicit integration schemes, achieving computational efficiency suitable for control applications. Building upon these foundations, the Geometric Variable Strain (GVS) approach}~\cite{mathew2025reduced} {represents the current state of the art in soft robot modeling, generalizing constant-curvature kinematics by parameterizing the continuous strain field of Cosserat rods using customizable basis functions, including polynomials, Legendre bases, or even Bezier curves. The GVS framework provides a unified Lagrangian formulation for both static and dynamic analysis of soft-rigid hybrid systems, and viscoelastic constitutive laws can be incorporated to capture rate-dependent material behavior. Della Santina et al.}~\cite{dellasantina2023soft} {provided a comprehensive structured taxonomy of these modeling approaches, systematically categorizing methods by their underlying mathematical assumptions and clarifying the theoretical relationships and trade-offs between model fidelity, computational cost, and generalizability.}

{Despite these theoretical advances, practical implementation of physics-based models for magnetically actuated soft surgical devices presents specific challenges. First, the GVS framework and similar approaches require accurate identification of constitutive parameters including Young's modulus, shear modulus, and damping coefficients. For novel 3D-printed silicone elastomers such as SIL~30, these parameters are difficult to characterize precisely, may vary with print orientation and cure conditions, and can exhibit batch-to-batch variability inherent in additive manufacturing. Second, magnetically actuated robots operating under rotating fields generated by multi-coil electromagnetic systems experience configuration-dependent torques where the magnetic loading changes as the device deforms, creating coupled magnetoelastic boundary conditions that must be explicitly formulated and updated during simulation. Third, the spatial variation of magnetic fields across the coil workspace introduces additional complexity for field mapping and real-time computation.}

{For instance, constant-curvature models assume uniform bending along discrete segments, yet magnetic actuation generates spatially distributed torques that vary with the local magnetic field orientation and gradient, resulting in non-uniform curvature profiles} \cite{Zhao_Ferromag_SciRob19, till2019real}. {Furthermore, soft silicone materials exhibit significant viscoelastic behavior that causes frequency-dependent deformation, and accurate knowledge of material parameters such as Young's modulus and damping coefficients is difficult to obtain for hyperelastic materials fabricated via 3D printing}~\cite{wang2021survey}. {The consequences of modeling errors are particularly critical for neurosurgical applications, where inaccurate shape prediction can lead to positioning errors at the device tip. Even millimeter-level deviations risk damaging delicate neurovascular structures such as the optic nerves, carotid arteries, or brainstem. Moreover, real-time closed-loop control, which is essential for safe surgical navigation in confined anatomical spaces, requires computationally efficient models that can be evaluated at high frequency.}}

{To address these practical challenges, recent studies have increasingly adopted data-driven and learning-based modeling strategies for soft and continuum robots. These frameworks use machine learning algorithms to learn the nonlinear mapping between actuation inputs and resulting deformation directly from experimental data, bypassing the need for explicit material characterization and constitutive model calibration. As discussed by Wang et al.}~\cite{wang2021survey}{, learning-based models address key challenges in soft robot modeling, including distributed compliance, material nonlinearity, and environmental interactions. Similarly, Falotico et al.}~\cite{falotico2025learning}{ demonstrated that learning controllers can adapt to dynamic conditions and fabrication variability while maintaining high accuracy and real-time computational performance.}

{Beyond these review articles, several primary studies have demonstrated the effectiveness of learning-based approaches for soft robot shape prediction and control. Almanzor et al.} \cite{almanzor2023static} {achieved static shape control of soft continuum robots using deep visual inverse kinematic models, showing that neural networks can effectively learn complex mappings between task-space targets and actuator inputs without explicit kinematic derivation. George Thuruthel et al.} \cite{georgethuruthel2024visuo} {introduced visuo-dynamic self-modeling for soft robotic systems, demonstrating that learning-based approaches can capture time-variant material properties and dynamic behaviors, including viscoelasticity and hysteresis, that are inherently difficult to incorporate in analytical models. These works provide compelling evidence that data-driven methods offer practical advantages for soft robot modeling, particularly when material properties are difficult to characterize or when complex dynamic effects dominate system behavior.}

{It is important to note that learning-based approaches are complementary to, rather than replacements for, physics-based frameworks such as GVS. While physics-based models offer superior interpretability, generalizability to untrained conditions, and capability for design optimization, learning-based methods provide a pragmatic pathway for rapidly characterizing device behavior when comprehensive material testing and magnetic field-body coupling formulation are not feasible. Future hybrid approaches that combine the GVS structure with learning-based parameter identification or model correction represent a promising research direction}~\cite{mathew2025reduced}. {Building upon these insights, this work extends the learning paradigm to magnetically actuated soft surgical instruments by introducing a data-driven framework that maps magnetic field parameters to geometric Bezier control points. {Unlike prior learning-based studies that primarily address tendon-driven or pneumatic soft robots, our framework specifically targets the unique challenges of magnetic actuation, where the loading depends on the spatial magnetic field distribution and the device's current configuration. To the best of our knowledge, this represents the first application of learning-based shape prediction specifically tailored for magnetically actuated soft surgical devices with embedded FBG sensing.} This compact parametric representation captures complex, spatially coupled effects such as magnetic torque distribution, viscoelastic response, and boundary constraints by training on experimental data collected across multiple actuation frequencies and field strengths.}

\begin{table}[!t]
\centering
\caption{{Geometric and material specifications of the magnetically steerable soft suction device.}}
\label{tab:device_specs}
\footnotesize
\setlength{\tabcolsep}{4pt}
\renewcommand{\arraystretch}{1.1}
\begin{tabular}{@{}l l@{}}
\toprule
\textbf{Parameter} & \textbf{Specification} \\
\midrule
Outer diameter & 4 mm \\
Inner suction channel & 2 mm diameter \\
Active flexible length & 40 mm \\
Tip magnet size & OD 6.35 mm, ID 3.18 mm, thickness 3.18 mm \\
Tip magnet magnetization & Through-thickness (axial) \\
Material & Biocompatible silicone elastomer (SIL 30) \\
Embedded sensor & Multi-core Fiber Bragg Grating (FBG) \\
FBG fiber outer diameter & 0.38 mm \\
Sensing length used & 40 mm (distal segment) \\
\bottomrule
\end{tabular}
\end{table}

With these challenges in mind, magnetic actuation has also emerged as a transformative solution for enhancing the control and flexibility of medical devices and robotic systems in minimally invasive procedures~\cite{yang2023magnetically, roshanfar2025advanced, roshanfar20253}. By applying external magnetic fields, forces and torques can be wirelessly transmitted to the distal end of instruments~\cite{abbott2020magnetic, he2025magnetically}, eliminating bulky mechanical linkages, cables, or fluidic systems. This approach reduces mechanical complexity, avoids issues like friction and backlash, and enables precise, remote control of devices within the body. Magnetic actuation has been successfully employed in medical applications such as catheter steering~\cite{dreyfus2024dexterous}, microsurgical tools~\cite{deng2024towards}, endoscopy~\cite{pittiglio2022patient}, targeted drug delivery~\cite{chen2022magnetically}, and microbiome sampling~\cite{lee2025magnetic}, among others. Typically, magnetically controlled robots are constructed either by attaching permanent magnets to the distal tip to create steerable catheters~\cite{Nelson_CardiacCatheter18}, or by embedding ferromagnetic particles within soft elastomers, such as PDMS or thermoplastic polyurethane, to fabricate flexible, continuum robots capable of controlled deformation~\cite{Zhao_Ferromag_SciRob19}. These designs allow safe navigation through intricate anatomical structures, improving surgical access, reducing patient trauma, and offering enhanced maneuverability~\cite{li2023dexterity, hong2020magnetic}.

Recognizing the potential of combining soft robotics and magnetic actuation, a prior study in~\cite{roshanfar2025soft} demonstrated the feasibility of integrating a magnetically steerable soft tip with a conventional rigid suction device to overcome the limitations of straight, rigid tools in endoscopic procedures. The prototype in that study featured soft segments made from Ecoflex 00-50 embedded with a permanent magnet, and it achieved controlled in-plane bending angles exceeding 90$^\circ$ under moderate magnetic fields (below 25~mT). The device was fabricated using mold-based casting techniques, offering both sufficient suction capability and flexibility. Validation experiments in simulated neurosurgical environments, using agar-based brain phantoms, demonstrated that the device could safely access target sites and perform precise tumor extraction, highlighting its potential to improve the safety and efficacy of minimally invasive neurosurgical interventions.

While the feasibility of the device has been previously studied, it relied on a simplified mechanical model based on constant curvature (CC) kinematics and an equivalent Young’s modulus to predict bending behavior~\cite{roshanfar2025soft}.
Although this approach provided reasonable accuracy for basic teleoperation, it was limited in its ability to fully capture the complex, nonlinear, and hyperelastic behavior of soft materials under varying magnetic fields and loading conditions. The CC assumption neglected strain-dependent stiffening, axial forces, shear effects, and dynamic interactions, particularly during large deformations or when subjected to non-uniform magnetic fields. Also, the fabrication process, based on mold-casting with non-biocompatible materials, constrained scalability, miniaturization, and clinical translation. To overcome these limitations (as summarized in the Table~\ref{tab:modeling}, this study introduces a novel magnetically steerable soft suction device designed specifically for endoscopic endonasal brain tumor resection, featuring a miniaturized, biocompatible structure fabricated via direct 3D printing. The device integrates embedded fiber Bragg grating (FBG) sensors to enable real-time shape sensing.

\begin{table*}[!t]
\setlength{\tabcolsep}{3pt}
\renewcommand{\arraystretch}{1.1}
\centering
\caption{Comparison of Analytical and Learning-Based Modeling Approaches in Soft Robotics.}
\label{tab:modeling}
\begin{tabularx}{\textwidth}{
  >{\raggedright\arraybackslash}p{2.1cm}
  >{\raggedright\arraybackslash}p{3.2cm}
  >{\raggedright\arraybackslash}p{2cm}
  >{\raggedright\arraybackslash}p{5.5cm}
  >{\raggedright\arraybackslash}p{5.5cm}
}
\hline
\textbf{Study} & \textbf{Application} & \textbf{Approach} & \textbf{Advantages} & \textbf{Limitations} \\
\hline
\makecell[lt]{Kim {\scriptsize\cite{Zhao_Ferromag_SciRob19}}} & Magnetic soft robots & Analytical & Magnetic continuum modeling & Static behavior only \\
\makecell[lt]{Roshanfar {\scriptsize\cite{roshanfar2023hyperelastic}}} & Hybrid soft robot & Analytical & Models hyperelastic stiffening & Not adaptive to dynamic inputs \\
\makecell[lt]{Masoumi {\scriptsize\cite{masoumi2024embedded}}} & Force sensing & Learning & Handles nonlinearities & Sensor-specific \\
\makecell[lt]{Torkaman {\scriptsize\cite{torkaman2023embedded}}} & 6-DoF sensing & Learning & Compact, embedded sensing & Hardware-specific model \\
\makecell[lt]{Pittiglio {\scriptsize\cite{pittiglio2022patient}}} & Magnetic catheter & Analytical & Autonomous, personalized motion & Requires imaging and maps \\
\makecell[lt]{He {\scriptsize\cite{he2025magnetically}}} & Neurosurgical tools & Analytical & Precise magnetic tip control & Rigid design; complex planning \\
\makecell[lt]{Dreyfus {\scriptsize\cite{dreyfus2024dexterous}}} & Endovascular robot & Analytical & High dexterity for vascular paths & No deformation modeling \\
\makecell[lt]{Hooshiar {\scriptsize\cite{hooshiar2020accurate}}} & Tip force estimation & Analytical & Sensor-free; physics-based & Requires exact motion input \\
\makecell[lt]{Jolaei {\scriptsize\cite{jolaei2020sensor}}} & Sensor-free control & Analytical & Lightweight contact model & Limited soft behavior accuracy \\
\makecell[lt]{Sayadi {\scriptsize\cite{sayadi2024finite}}} & Shape prediction & Analytical & Extended piecewise constant curvature & Requires segmentation \\
\makecell[lt]{{Santina {\scriptsize\cite{dellasantina2023soft}}}} & {Soft robot modeling} & {Analytical} & {Structured taxonomy; comprehensive framework comparison} & {Review/taxonomy paper} \\
\makecell[lt]{{Mathew} {\scriptsize\cite{mathew2025reduced}}} & {Hybrid soft-rigid robots} & {Analytical} & {Geometrically exact; customizable strain bases; unified dynamic framework} & {Requires material parameter identification} \\
\makecell[lt]{{Almanzor {\scriptsize\cite{almanzor2023static}}}} & {Shape control} & {Learning} & {Deep visual inverse kinematics; task-space control} & {Requires visual feedback} \\
\makecell[lt]{{Thuruthel {\scriptsize\cite{georgethuruthel2024visuo}}}} & {Dynamic modeling} & {Learning} & {Visuo-dynamic self-modeling; captures time-variant properties} & {Vision-based; system-specific} \\
\makecell[lt]{\textbf{This work}} & Magnetic soft robot & Learning & Real-time, accurate and robust & Needs experimental dataset \\
\hline
\end{tabularx}
\end{table*}

Shape estimation in flexible robots can generally be achieved using either sensor-free or sensor-based methods. Sensor-free approaches rely on kinematic or magnetic models to infer shape without embedded sensors \cite{jolaei2020sensor, roshanfar2023hyperelastic}, while sensor-based techniques use embedded elements for direct measurement \cite{masoumi2024embedded, torkaman2023embedded}. Our method follows the sensor-based approach, combining FBG data with a learning-based model to enable accurate, real-time shape prediction. {More importantly, a learning-based modeling framework is proposed to replace the traditional mechanical models. The proposed framework directly maps magnetic field inputs to the device’s geometric configuration using Bezier control points, providing a compact and interpretable representation of the robot shape. This data-driven approach eliminates the need for explicit mechanical assumptions and instead leverages experimental data to capture the coupled nonlinear behavior between magnetic actuation and hyperelastic deformation. By combining physical sensing through embedded FBGs with data-driven modeling, the framework achieves accurate and real-time shape prediction suitable for closed-loop surgical control applications. This enables precise shape prediction, improved control, and adaptability across diverse surgical scenarios. To the best of our knowledge, this is the first magnetically actuated soft suction device equipped with a learning-based shape prediction model tailored for neurosurgical applications}. The key contributions are:
\begin{itemize}
    \item Design and fabrication of a miniaturized, 3D-printed magnetically steerable soft suction device using biocompatible material, with integrated FBG sensors for real-time shape sensing,
    \item {Introduction of a novel learning-based modeling paradigm that connects magnetic actuation inputs directly to Bezier control points, providing a compact, interpretable, and low-dimensional representation of the device geometry suitable for real-time control,}
    \item {Conceptual integration of magnetic field characterization, embedded FBG shape sensing, and Bezier-based learning into a unified modeling framework that can be generalized to other magnetically actuated soft robotic systems,}
    \item Experimental validation demonstrating high shape prediction accuracy across a range of magnetic field strengths and actuation frequencies, effectively addressing the nonlinear behavior of hyperelastic materials. 
\end{itemize}

\section{Materials and Methods}
\label{Materials and Methods}
\subsection{Shape Encoding}
\label{Shape Encoding}
Accurate shape reconstruction is essential for controlling flexible, magnetically actuated soft robots. In this work, shape sensing is achieved using a multi-core fiber (MCF) embedded with fiber Bragg gratings (FBG), combined with a Bezier curve-based representation to provide a compact and smooth model of the device's deformation. This approach not only reduces high-dimensional sensor data into a small set of control points but also ensures smoothness and continuity in the reconstructed shape. The resulting representation is well-suited for integration into learning-based models, enabling both efficient training and real-time inference.

\subsubsection{FBG-Based Shape Sensing with Multi-Core Fiber}
\label{FBG-Based Shape Sensing with Multi-Core Fiber}
The soft suction device was integrated with an MCF containing 26 FBGs uniformly spaced at 10~mm intervals along a sensing length of 250~mm. The MCF is an off-the-shelf shape-sensing sensor purchased from FBGS company (FBGS, Belgium). The outer diameter (OD) of the FBG fiber was 0.38~mm, allowing seamless embedding within the inner channel of the soft robot. FBGs function by reflecting a specific wavelength of light, known as the Bragg wavelength ($\lambda_B$), which shifts in response to axial strain ($\Delta \varepsilon$) and temperature variations ($\Delta T$). This shift is governed by~\cite{al2021fbg}:
\begin{equation}
\frac{\Delta \lambda_B}{\lambda_{B0}} = S_{\varepsilon} \Delta \varepsilon + S_T \Delta T
\label{eq:bragg_shift}
\end{equation}
where $\lambda_{B0}$ is the initial Bragg wavelength, $S_{\varepsilon}$ and $S_T$ are sensitivity coefficients for strain and temperature, respectively. In the symmetric MCF configuration, the central core serves for temperature compensation, while the outer cores measure bending-induced strain. The strain in each outer core is:
\begin{equation}
\Delta \varepsilon_i = \frac{\Delta \lambda_{B,i}}{S_{\varepsilon} \lambda_{B0,i}} - \frac{\Delta \lambda_{B,1}}{S_{\varepsilon} \lambda_{B0,1}}
\label{eq:strain_calc}
\end{equation}
then, the curvature ($\kappa$) and bending angle ($\theta_b$) at each grating location are derived using \cite{al2021fbg}:
\begin{equation}
\kappa = \frac{2 \left\| \boldsymbol{\kappa}_{\text{app}} \right\|}{N}, \quad\quad \theta_b = \angle \boldsymbol{\kappa}_{\text{app}}
\end{equation}
where:
\begin{equation}
\boldsymbol{\kappa}_{\text{app}} = -\sum_{i=1}^{N} \frac{\varepsilon_i}{r} \cos(\theta_i) \, \hat{\mathbf{i}} - \sum_{i=1}^{N} \frac{\varepsilon_i}{r} \sin(\theta_i) \, \hat{\mathbf{j}}
\end{equation}
and:
\begin{equation}
\varepsilon_i = -\kappa r \sin\left(\theta_b - \frac{3\pi}{2} - \theta_i\right)
\label{varepsilon}
\end{equation}
where $r$ is the radial distance from the center to the outer cores (one central core with three evenly spaced radial cores), $N=3$ is the number of outer cores, and $\theta_i$ is the angular position of each core. The curvature vector $\boldsymbol{\kappa}_{\text{app}}$ denotes the apparent bending curvature at each segment, expressed as a 2D vector in the plane orthogonal to the fiber axis. Its magnitude represents the bending curvature, while its angle defines the bending plane direction $\theta_b$. Note that this formulation assumes planar bending between FBG nodes and does not account for geodesic curvature or torsion.

Next, the MCF embedded within the soft robot was connected to an optical interrogator (FBGS, Belgium), which continuously monitored wavelength shifts in each of the 26 FBGs. These wavelength shifts, induced by bending-related strain, were sampled in real-time (30 Hz) and transmitted to a processing unit via a transmission control protocol (TCP) connection established with a custom MATLAB script. For each grating, the interrogator provided axial strain values from the outer cores of the MCF. Using Eq.~\ref{eq:strain_calc}, temperature-compensated strain $\Delta \varepsilon_i$ was calculated by referencing the central core, which is insensitive to bending. The resulting strain values were then converted into local curvature ($\kappa$) and bending plane angle ($\theta_b$) for each grating position, following the geometric relationship defined in Eq.~\ref{varepsilon}.

Once the curvature and curvature bending angle profiles were obtained, the MATLAB script performed continuous shape reconstruction of the soft robot's 3D centerline by applying piecewise constant curvature (PCC) interpolation between each grating. The total sensing length was discretized into 25 uniform segments, each corresponding to the interval between adjacent FBGs with a fixed arc length of 10~mm. For each segment, if the curvature $\kappa$ was close to zero, it was treated as a straight section with a pure translation along the local axial direction. When curvature was present, the segment was modeled as a circular arc, where the radius was defined as $r = 1/\kappa$ and the bending angle $\theta$ was computed as $\kappa \cdot ds$. A local rotation matrix was constructed for each segment to align the bending direction according to $\theta_b$, and the corresponding transformation was applied. These local transformations, consisting of rotation and translation, were sequentially propagated along the length of the fiber to build the global shape. 

To mitigate numerical drift accumulated through successive matrix multiplications, singular value decomposition (SVD) was applied at each step to maintain orthogonality of the rotation matrices. The 3D Cartesian coordinates of the robot’s centerline were extracted from the origins of the cumulative transformation matrices, resulting in a set of 26 points representing the reconstructed backbone. Finally, the total arc length of the reconstructed shape was computed to ensure that the final shape accurately matched the known physical length of 250~mm. This reconstruction process effectively converts raw optical strain data into a precise 3D representation of the soft robot’s shape. By leveraging the PCC assumption, the algorithm achieves a balance between computational efficiency and sufficient accuracy for real-time surgical navigation. The reconstructed centerline serves as the basis for fitting a Bezier curve, providing a compact and smooth representation suitable for integration into the learning-based control framework.
{It is important to note that this reconstruction step employs {continuous shape reconstruction} to obtain a smooth 3D centerline from discrete FBG curvature measurements{, implemented via piecewise constant curvature (PCC) interpolation between grating nodes, which is a standard practice in FBG-based shape sensing}~\cite{al2021fbg, dong2022shape}. {This is fundamentally different from the ``traditional PCC-based modeling'' discussed in the Introduction, which refers to analytical kinematic models that derive closed-form actuation-to-shape relationships under idealized constant-curvature assumptions. Such analytical models fail to capture spatially distributed magnetic torques, material viscoelasticity, and dynamic coupling effects.}
{In contrast, our learning-based framework imposes no constant-curvature constraint on the predicted shapes.} The Bezier curve fitting applied in the following section provides a smooth and compact geometric representation that abstracts the overall deformation into a small number of control points. These control points are then used as learning targets, {encoding the experimentally measured deformation under each actuation condition. By training on data collected across multiple frequencies (0.2--1.0~Hz) and magnetic field strengths (0--14~mT), the model learns the actual nonlinear, frequency-dependent, and viscoelastic behavior of the soft robot, effects that are inherently captured in the experimental measurements but would be absent in analytical PCC-based kinematic models. We acknowledge that the ground truth accuracy is bounded by the FBG reconstruction fidelity; however, the learning model's ability to capture dynamic and coupled effects from experimental data represents a fundamental advantage over traditional analytical approaches.}}

\subsubsection{Bezier Curve Representation of the Soft Robot Shape}
\label{Bezier Curve Representation of the Soft Robot Shape}
Flexible continuum robots, such as the magnetically actuated soft suction device in this study, naturally exhibit smooth and continuous deformation patterns. These characteristics make them ideal candidates for representation using spline-based methods. Among various spline families, Bezier curves are widely adopted due to their mathematical simplicity, ability to generate smooth curves, and intuitive geometric control through a limited number of points~\cite{hooshiar2020accurate, sayadi2024finite}. A cubic Bezier curve, in particular, offers the minimal degree necessary to approximate constant-curvature shapes, which are common in soft robotic and catheter-like structures. In this study, a cubic Bezier curve was employed to approximate the shape of the soft robot reconstructed from discrete FBG sensor data. The Bezier curve defines a continuous path, parameterized by $s \in [0,1]$, using four control points $\mathbf{p}_0$, $\mathbf{p}_1$, $\mathbf{p}_2$, and $\mathbf{p}_3$, as:
\begin{equation}
\mathbf{c}(s) = (1-s)^3 \mathbf{p}_0 + 3s(1-s)^2 \mathbf{p}_1 + 3s^2(1-s) \mathbf{p}_2 + s^3 \mathbf{p}_3
\label{eq:Bezier_curve}
\end{equation}
where, $\mathbf{c}(s)$ represents the position vector along the curve at any normalized parameter $s \in [0,1]$, where $s = 0$ indicates the base of the curve ($\mathbf{p}_0$) and $s = 1$ indicates the tip of the curve ($\mathbf{p}_3$). The first and last control points, $\mathbf{p}_0$ and $\mathbf{p}_3$, are constrained to coincide exactly with the base and tip points of the FBG-reconstructed centerline, ensuring that the Bezier curve satisfies the physical boundary conditions of the soft robot in each frame. The intermediate control points, $\mathbf{p}_1 $ and $\mathbf{p}_2$, govern the curvature profile and the smooth transition along the curve. An important modeling assumption is that the total length of the soft robot remains constant during magnetic actuation. This constraint ensures that the Bezier curve fitting adheres to the physical behavior of the device without introducing artificial elongation or compression. The role of $\mathbf{p}_1$ and $\mathbf{p}_2 $ extends beyond simple geometric placement. As highlighted in~\cite{hooshiar2020accurate}, the positioning of these control points directly influences the curvature characteristics of the Bezier curve. A shape parameter $\lambda$ can be introduced to conceptually control the convexity of the curve. While in this study $\mathbf{p}_1$ and $\mathbf{p}_2$ were determined through a least-squares fitting process, understanding the influence of $\lambda$ provides valuable insight into how control point spacing affects bending behavior. Smaller values of $\lambda$ correspond to tighter curvature, while larger values result in smoother, more gradual bends.

\begin{figure}[!t]
    \centering
    \includegraphics[width=\linewidth]{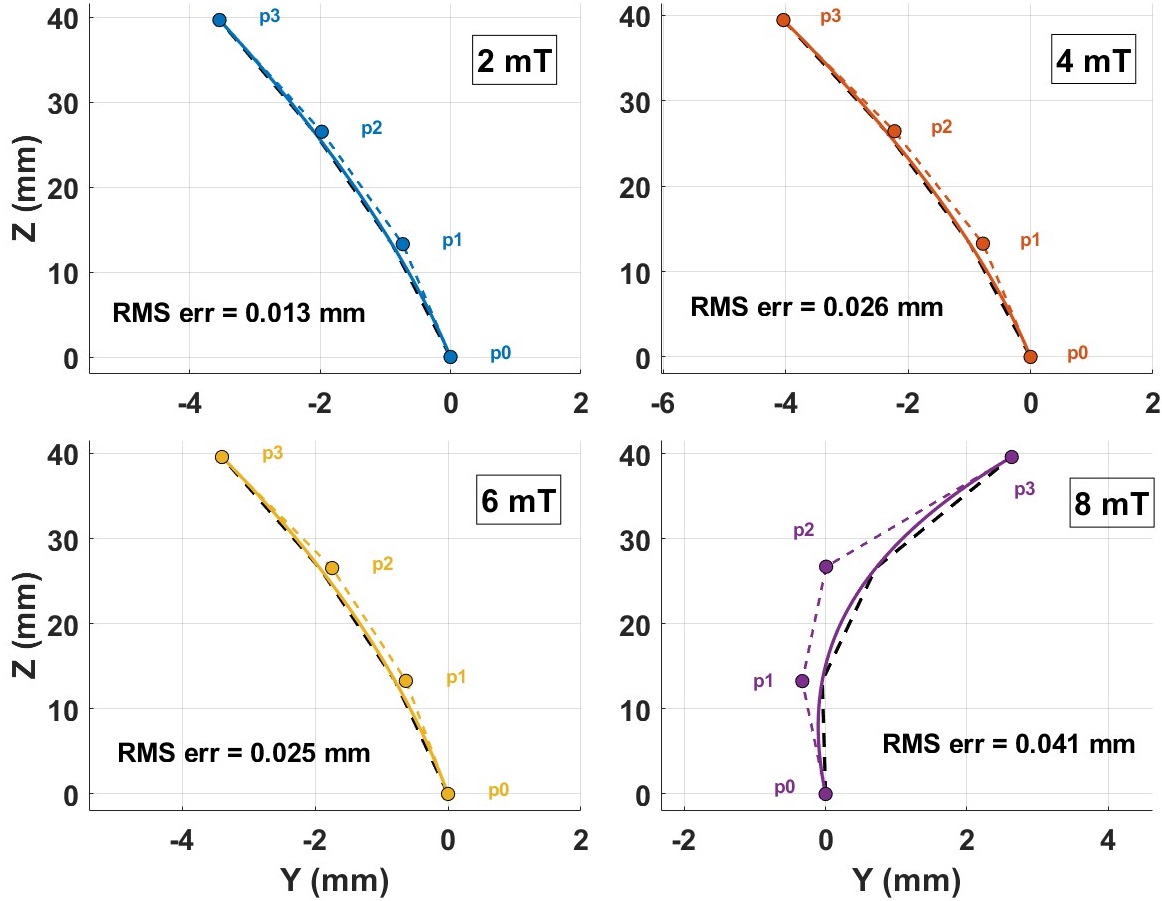}
    \caption{FBG-reconstructed shapes (black dashed lines) and corresponding Bezier curve fits (colored solid lines) for four representative frames during 1.0 Hz magnetic field rotation with the soft robot tip positioned 90~mm above the coil table.}
    \label{fig:Bezier_fit}
\end{figure}

Using a Bezier curve to represent the shape offers substantial advantages over directly handling the full set of FBG-derived points. Reducing the description from 26 discrete points to just four control points significantly lowers the dimensionality of the data. This compact representation facilitates faster processing, reduces memory requirements, and is particularly advantageous in learning-based models where lower-dimensional outputs improve training efficiency and real-time inference. Furthermore, Bezier curves inherently provide smooth and continuous approximations, which effectively filter out noise present in sensor data, yielding a clearer and more reliable representation of the robot's deformation.
The fitting process was formulated as a least-squares optimization problem, aiming to minimize the cumulative Euclidean distance between the continuous Bezier curve $\mathbf{c}(s)$ and the discrete centerline points obtained from FBG reconstruction. Since the number of data points exceeds the number of parameters to estimate, this creates an overdetermined system. The Moore--Penrose pseudoinverse was used to compute the unique solution that minimizes the total squared error, ensuring a stable and balanced fit even in the presence of sensor noise or minor inconsistencies in the data. Let $\mathbf{c}(s_i) \in \mathbb{R}^3$ denote the reconstructed centerline points at normalized arc-length parameters $s_i \in [0,1]$. Since the endpoints $\mathbf{p}_0$ and $\mathbf{p}_3$ are fixed to match the FBG base and tip positions, the problem reduces to solving for the intermediate control points $\mathbf{p}_1$ and $\mathbf{p}_2$ that minimize the fitting error:
\begin{equation}
\min_{\mathbf{p}_1, \mathbf{p}_2} \sum_{i=1}^{N} \left\| \mathbf{c}(s_i) - \mathbf{Q}(s_i) 
\begin{bmatrix} \mathbf{p}_1 \\ \mathbf{p}_2 \end{bmatrix} \right\|^2
\end{equation}
where $\mathbf{Q}(s_i) \in \mathbb{R}^{3 \times 6}$ is the Bezier basis matrix for $\mathbf{p}_1$ and $\mathbf{p}_2$ at $s_i$. Then, the optimal solution is:
\begin{equation}
\begin{bmatrix} \mathbf{p}_1 \\ \mathbf{p}_2 \end{bmatrix} 
= \left( \mathbf{Q}^\top \mathbf{Q} \right)^{\dagger} \mathbf{Q}^\top \mathbf{C}
\end{equation}
where $\mathbf{C} = [\mathbf{c}(s_1)^\top, \dots, \mathbf{c}(s_N)^\top]^\top \in \mathbb{R}^{3N}$ is the stacked vector of FBG-derived centerline coordinates. By leveraging fundamental properties of Bezier curves, such as $\mathbf{c}(0) = \mathbf{p}_0$ and $\mathbf{c}(1) = \mathbf{p}_3$, the method inherently satisfies positional boundary conditions. The arc-length constraint is enforced during post-processing by scaling the fitted curve to match the known physical length of the device. Figure~\ref{fig:Bezier_fit} illustrates a representative experimental example (for the case of 1~Hz rotation and a 90~mm distance between the tip and the coil table surface) of the FBG-reconstructed shape alongside the fitted Bezier curve and its control points. The fitting RMSE between the FBG-reconstructed shapes and the corresponding Bezier curves is also reported for these representative cases. This confirms that Bezier curves provide a computationally efficient, compact, and accurate representation of soft robot deformation, making them highly suitable for integration into the learning-based modeling framework proposed in this study.

\subsection{{Experimental Setup}}
\label{Experimental Setup}
\begin{figure*}[!t]
    \centering
    \includegraphics[width=\linewidth]{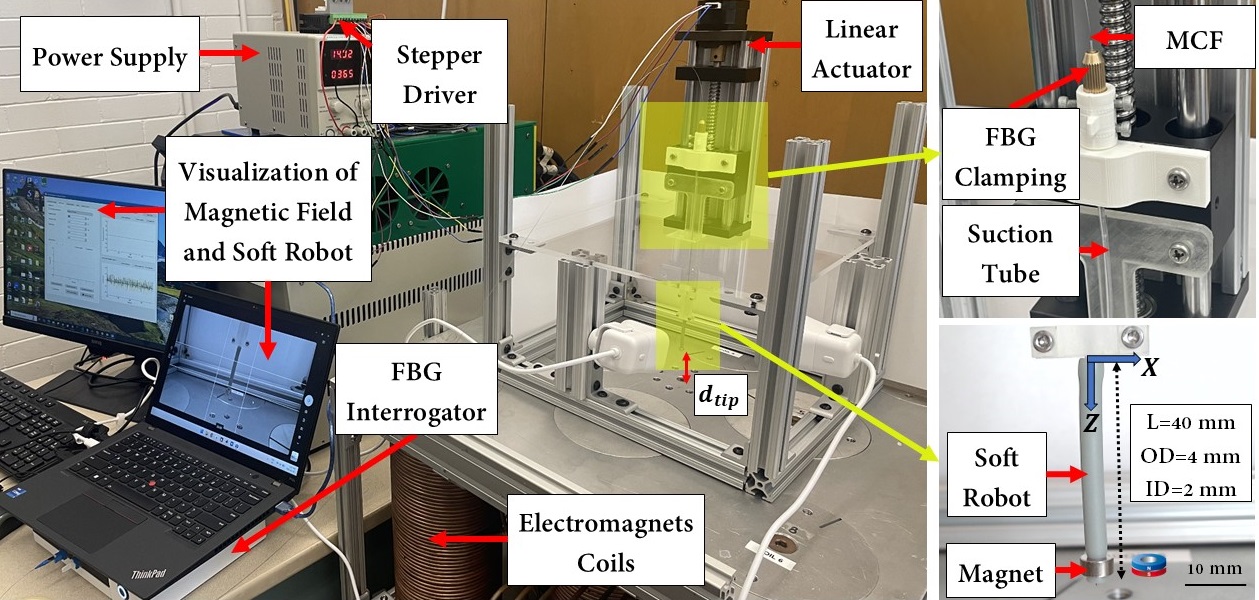}
    \caption{Experimental setup for magnetic actuation and shape sensing of soft robot. The system includes an FBG-embedded soft robot mounted on a linear actuator, positioned above electromagnetic coils at a distance of $d_{tip}$. A power supply and stepper driver control the linear actuation, while real-time strain data is collected via the FBG interrogator and visualized on a workstation. Close-ups on the right show the FBG clamping, suction tube, 3D printed fixtures, and the soft robot with a tip magnet for steering.}
    \label{fig:setup}
\end{figure*}

The experimental setup was designed to evaluate the deformation behavior of the soft robot under controlled magnetic fields while capturing real-time shape data through embedded FBG sensors. The soft suction device was fabricated using biocompatible SIL 30 material (Carbon Inc., Redwood City, California, USA) via digital light synthesis (DLS) 3D printing, ensuring precise geometry and material consistency suitable for medical applications. SIL 30 is a silicone urethane elastomer with a nominal Shore A hardness of 35 (instant), as reported by the manufacturer~\cite{carbon_sil30_tds}. The device measured 40~mm in length, with an OD of 4~mm and an inner channel diameter of 2~mm. A hollow cylindrical neodymium permanent magnet (Grade N50, 6.35~mm OD $\times$ 3.18~mm ID $\times$ 3.18~mm thickness, nickel-coated) was securely affixed to the distal tip of the device to enable magnetic steering. The magnet features through-thickness magnetization as shown in Fig.~\ref{fig:setup}, optimized for effective actuation within rotating magnetic fields.

The FBG fiber was threaded through the suction channel to enable continuous shape sensing along the length of the soft robot. The proximal end of the fiber was fixed at the base, while the distal end was anchored at the robot’s tip to ensure consistent alignment and accurate strain transfer during deformation. Owing to the fiber’s small OD (0.38~mm), its integration within the lumen imposed negligible mechanical constraint on the flexibility and natural deformation of the soft robot. The proximal end of the MCF was clamped onto a linear actuator, which allowed fine control over the vertical positioning of the soft robot relative to the electromagnetic coil system's surface. Both the soft robot and the linear actuator were mounted onto a custom-designed aluminum frame to maintain structural stability and ensure consistent experimental conditions throughout data collection. Additionally, two 3D-printed supports were employed to secure both the base of the FBG fiber and the soft robot to the linear actuator, minimizing unwanted vibrations or displacement. For visual monitoring, two Logitech cameras were positioned orthogonally, providing multi-angle views of the robot’s deformation during actuation.
\begin{figure}[!t]
  \centering
  \includegraphics[width=\linewidth]{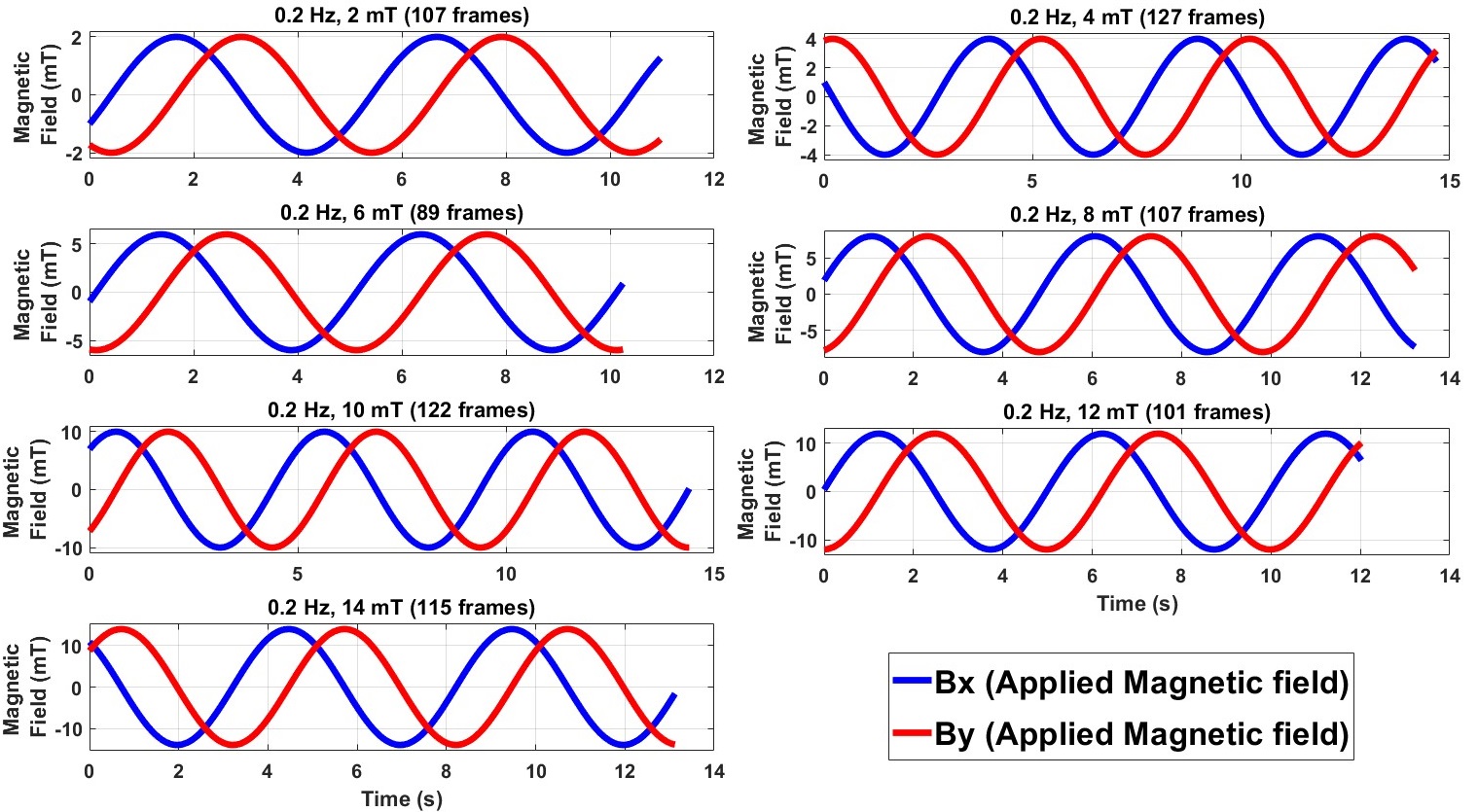}
  \caption{Time-varying magnetic field components ($b_x$, $b_y$) for the 0.2~Hz trials corresponding to the trajectories in Fig.~\ref{fig:shape3d}.}
  \label{fig:field_profiles}
\end{figure}

\subsection{{Soft Robot Characterization}}
\label{Soft Robot Characterization}
\begin{figure}[!t]
 \centering
  \includegraphics[width=0.95\linewidth]{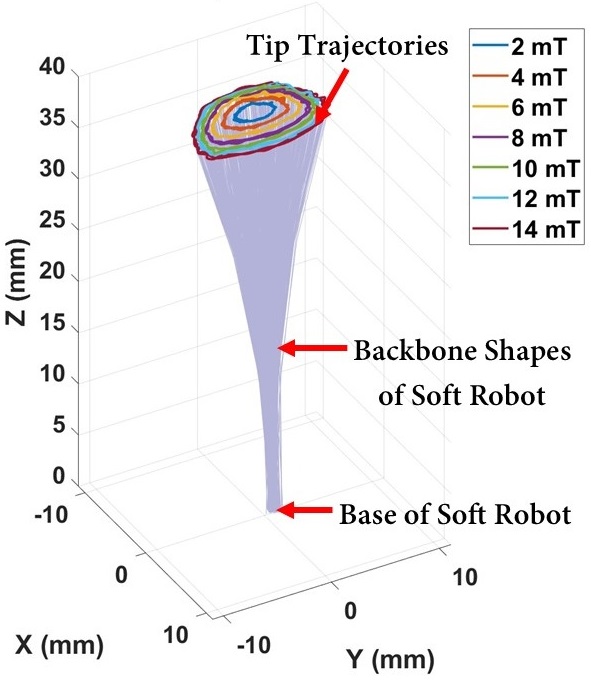}
  \caption{3D shape reconstruction (0.2 Hz). Each curve represents a centerline shape reconstructed from FBG data at different magnetic field strengths.}
 \label{fig:shape3d}
\end{figure}

Magnetic actuation was achieved using an electromagnetic navigation system comprising eight high-power coils arranged beneath the workspace~\cite{schonewille2024electromagnets}. This system was engineered to deliver spatially controllable magnetic fields while maximizing workspace accessibility. The coil system was capable of generating magnetic fields up to 38~mT along the $x$ and $y$ axes, and up to 47~mT along the $z$ axis at a working height of 120~mm above the coil surface. During each experimental run, an in-plane rotating magnetic field was applied parallel to the $xy$ plane to induce controlled bending and rotation of the robot’s tip (Fig.~\ref{fig:field_profiles}). The applied magnetic field vector followed a sinusoidal form given by:
\begin{equation}
\mathbf{b}(t) = A[-\sin(\omega t),\ \cos(\omega t),\ 0]^\top
\end{equation}
where $A$ is the magnetic field magnitude (ranging from 0 to 14~mT) and $\omega = 2\pi f$ is the angular frequency corresponding to the rotation frequency $f$ (0.2, 0.4, 0.6, 0.8, and 1~Hz). This dynamic formulation generated a uniform circular magnetic field in the $xy$ plane, producing a time-varying torque on the embedded tip magnet that resulted in cyclic bending motion. 
As shown in Fig.~\ref{fig:field_profiles}, the blue and red curves represent the $b_x$ and $b_y$ components, respectively, generating a circular rotating field in the XY plane.
To illustrate the magnetic field conditions applied during soft robot actuation, we visualized the in-plane rotating magnetic field components ($b_x$, $b_y$) for the 0.2 Hz excitation frequency across magnetic field magnitudes from 2 mT to 14 mT ($b_z=0$). These sinusoidal field profiles correspond directly to the workspace visualizations presented in Fig.\ref{fig:shape3d}, where tip deformation was analyzed under identical field conditions. 

\begin{figure}[!t]
  \centering
  \subfloat[XY View (0.2 Hz)]{
    \includegraphics[width=0.49\linewidth]{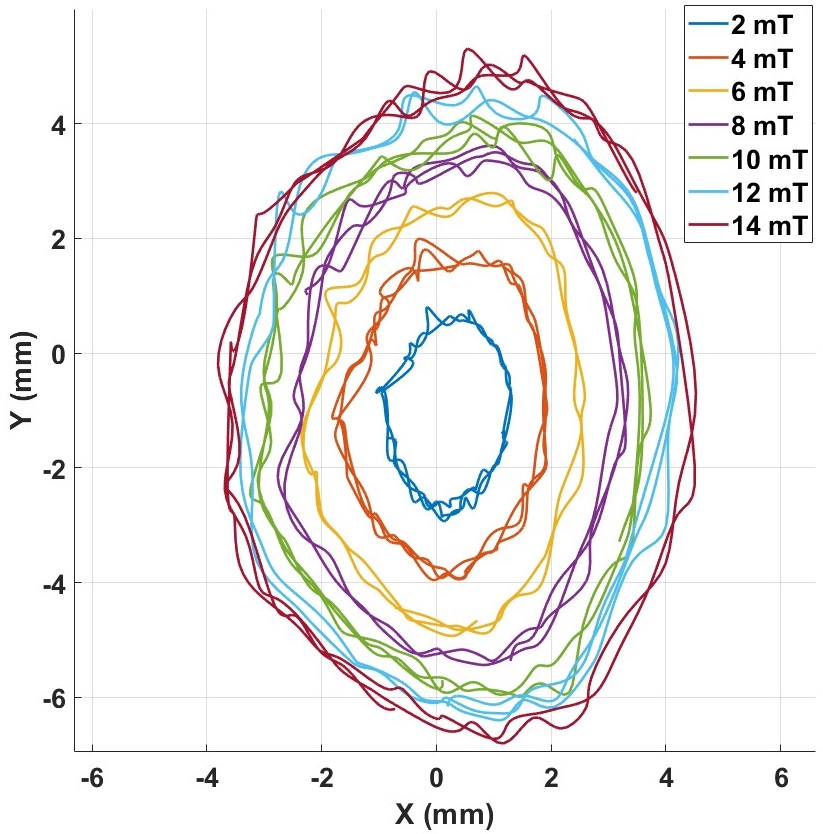}}
  \hfill
  \subfloat[XY View (0.4 Hz)]{
  \includegraphics[width=0.49\linewidth]{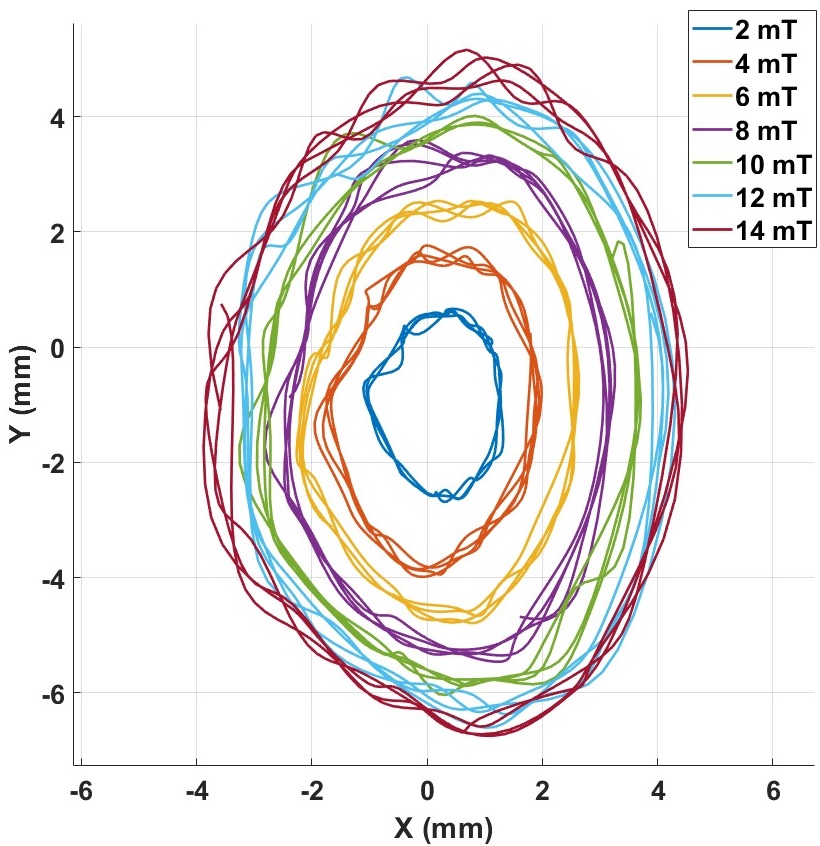}}
  \vspace{1ex}
  \subfloat[XY View (0.6 Hz)]{
  \includegraphics[width=0.49\linewidth]{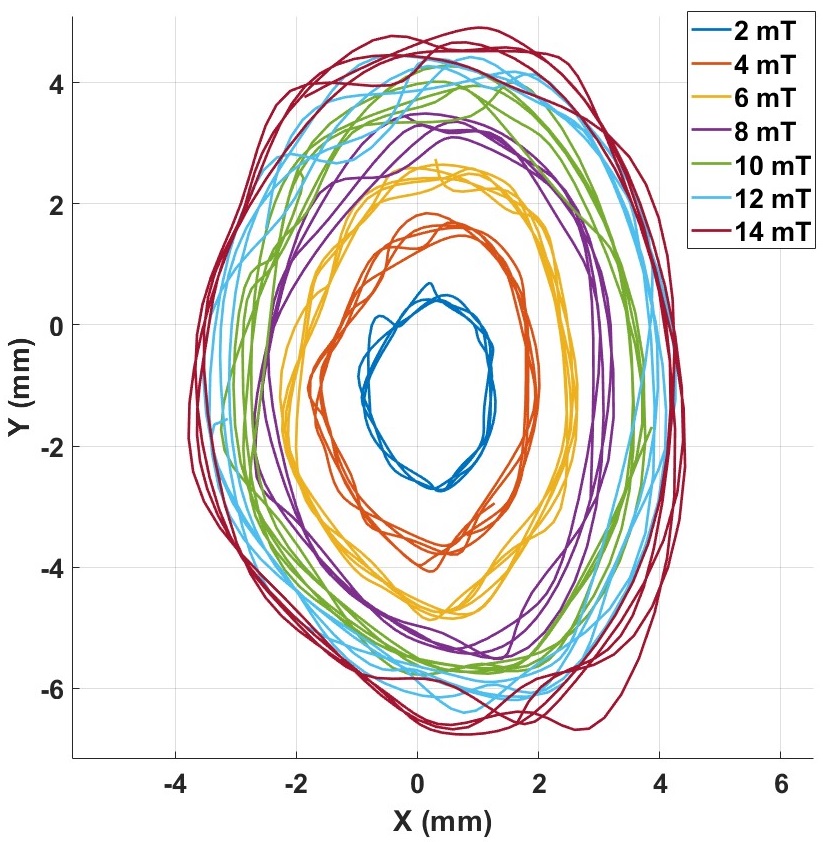}}
  \hfill
  \subfloat[XY View (0.8 Hz)]{
  \includegraphics[width=0.49\linewidth]{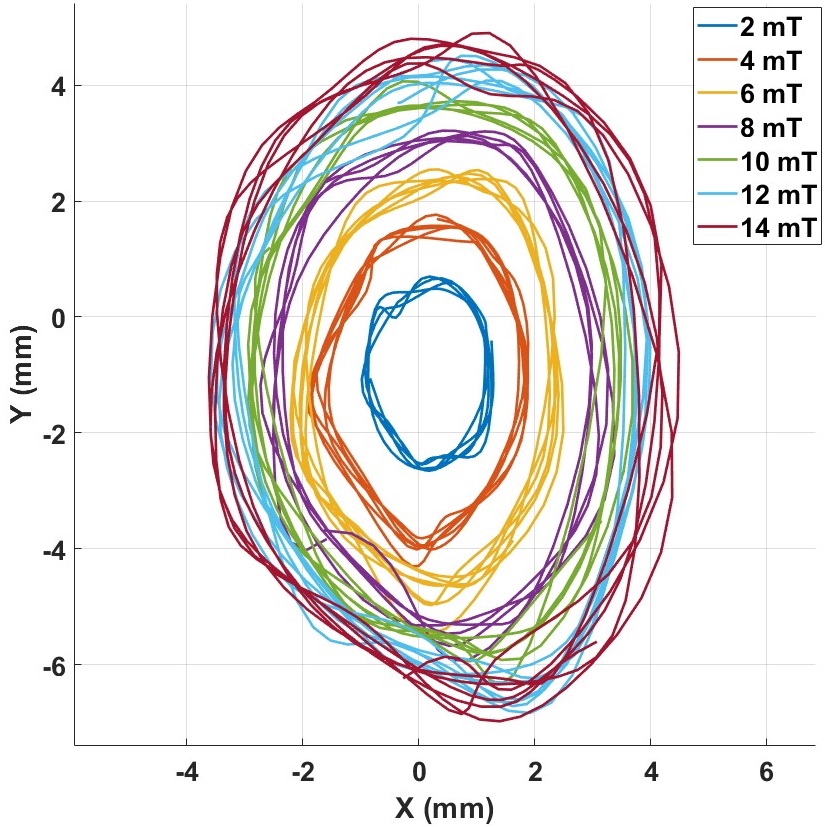}}
  \vspace{1ex}
  \subfloat[XY View (1.0 Hz)]{
  \includegraphics[width=0.49\linewidth]{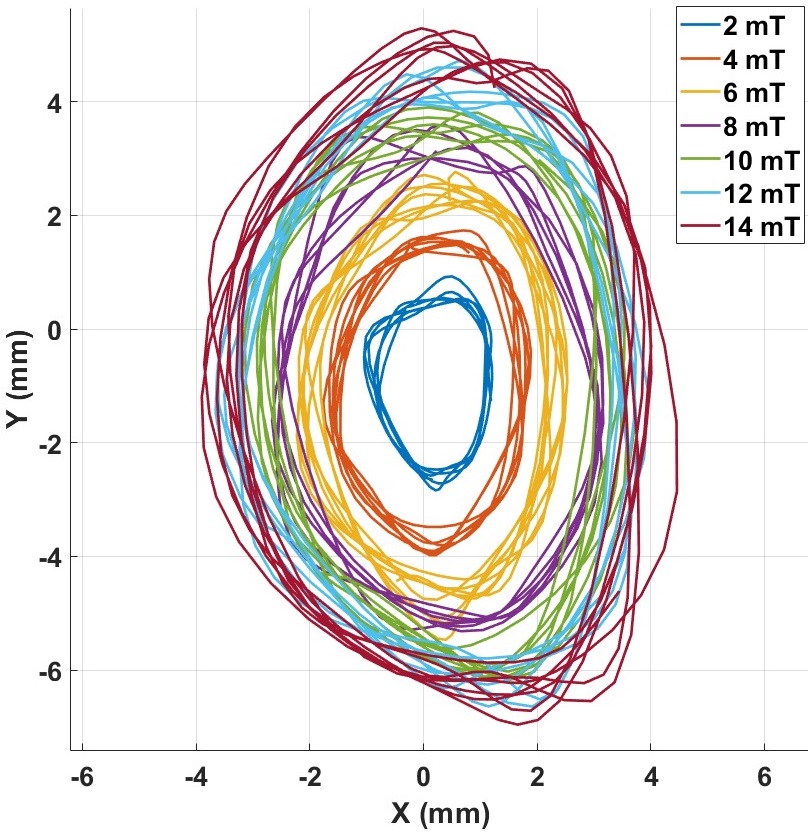}}
  \caption{XY-view trajectories at five actuation frequencies. Each panel shows tip paths reconstructed from FBG data under varying magnetic field strengths.}
  \label{fig:xy_trajs}
\end{figure}

A dynamic field was deliberately applied to capture the frequency dependent deformation characteristics of the soft robot, which are influenced by the viscoelastic behavior of the silicone material. As shown in Figs.~\ref{fig:xy_trajs}, the same magnetic field magnitude led to larger rotation radii at higher frequencies, highlighting the presence of dynamic effects. The chosen frequency range (0.2–1~Hz) was selected to reflect typical motion speeds observed during endoscopic procedures, ensuring the evaluation remained within clinically relevant operational conditions. To record the data, the robot’s tip was initially positioned 100~mm above the electromagnetic coil table surface. Upon completion of the test series at this height, the linear actuator was used to adjust the tip position to 90~mm above the table surface, and the full set of experiments was repeated under identical conditions to assess the influence of vertical distance on soft robot performance. These two heights were selected to represent clinically relevant working distances that fall within the safe operating range of the electromagnetic coil system, while ensuring measurable differences in magnetic field strength and resulting deformation. In total, 70 distinct actuation scenarios were conducted, covering all combinations of magnetic field strengths, rotation frequencies, and the two vertical distances. Throughout these tests, the strain data from all 26 FBG gratings were continuously recorded at a sampling rate of 30~Hz using a custom MATLAB script interfaced via TCP with the optical interrogator. Following data acquisition, the recorded 250~mm sensing length of the MCF was processed in post-experiment analysis. Only the final 40~mm segment, which corresponds precisely to the embedded length within the soft robot, was retained for shape reconstruction and modeling.

\section{{Results and Discussion}}
\label{EVALUATIONS AND DISCUSSION}
{This section presents a comprehensive evaluation of the learning based approaches used for modeling the soft suction device. The comparison between the Neural Network (NN) and Random Forest (RF) models was selected to represent two complementary learning paradigms for nonlinear regression. The NN provides a parametric approach suitable for learning smooth mappings between magnetic field inputs and geometric outputs, while the RF offers a non-parametric ensemble method that is more robust to noise and better suited for moderate-sized experimental datasets. Since the dataset used in this study consists of 5,097 experimentally collected samples with inherent variability and measurement noise, comparing these two models enables assessment of their relative accuracy, stability, and computational efficiency for predicting Bezier control points in real time. We compare the performance of both models in predicting Bezier control points from magnetic field parameters, analyze feature importance, and evaluate shape reconstruction accuracy in the following sections.}

\subsection{Dataset Characteristics and Model Training}
\label{Dataset Characteristics and Model Training}
\begin{figure}[!t]
   \centering
   \includegraphics[width=\linewidth]{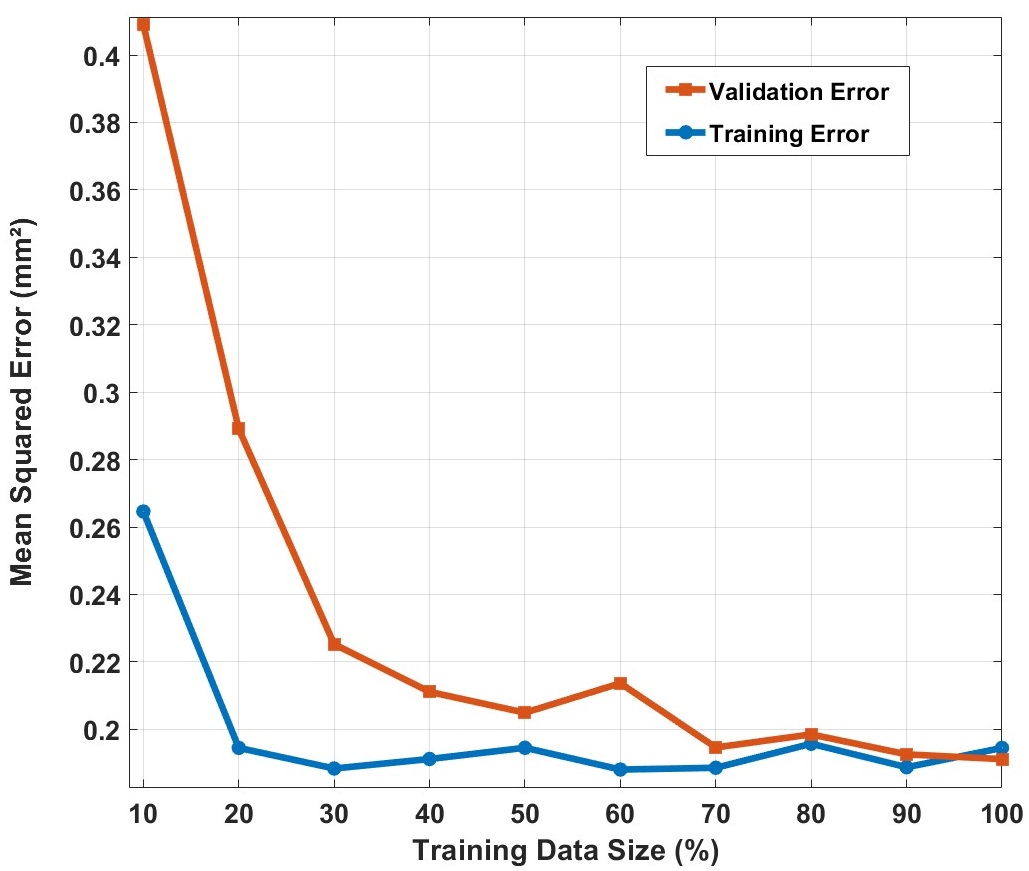}\vspace{5mm}
   \includegraphics[width=\linewidth]{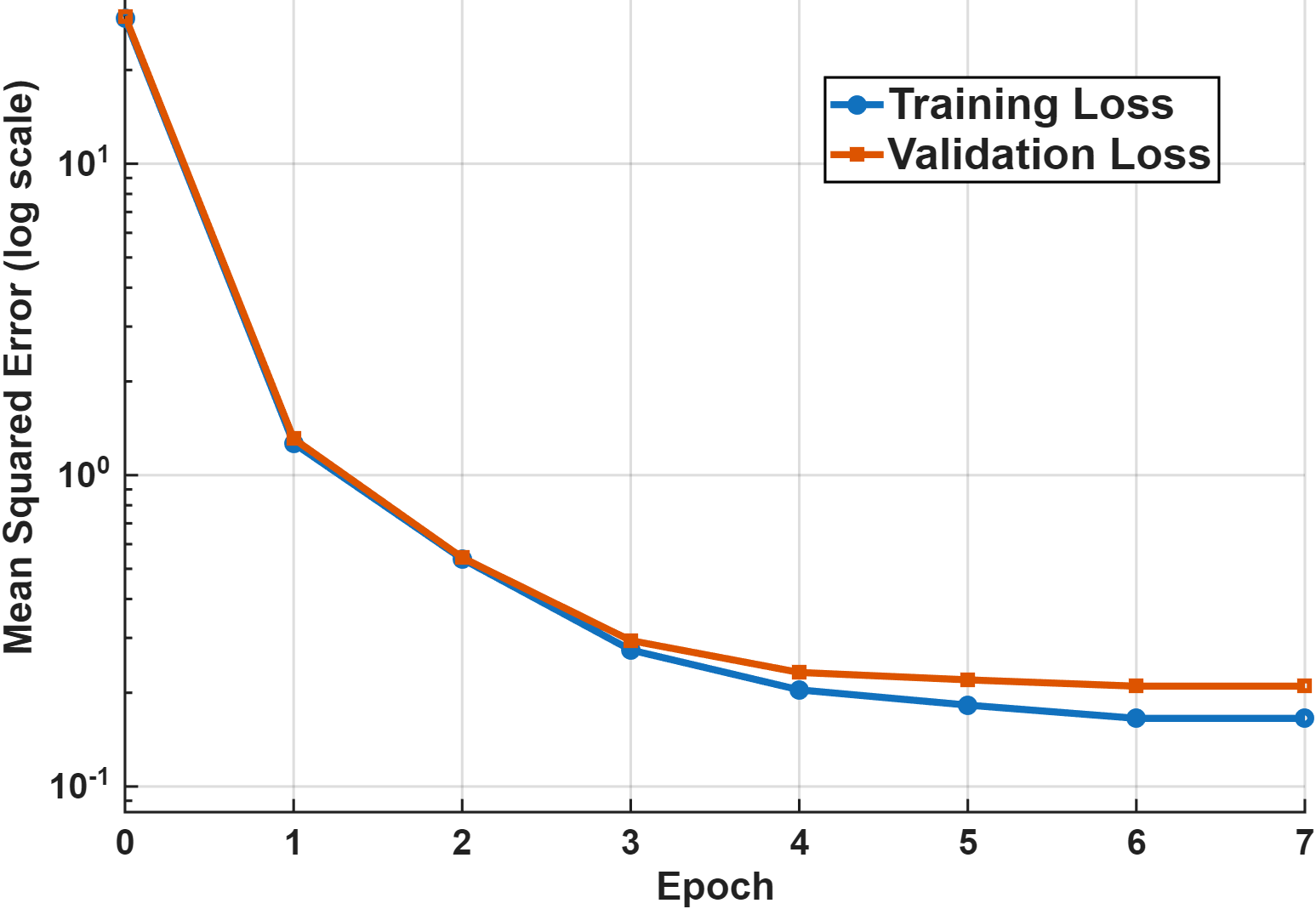}
   \caption{Neural network training diagnostics.
   (Top) Learning-curve analysis showing mean squared error versus training data size. Performance improves quickly up to roughly 30\% of the data and then levels off, indicating efficient learning from a relatively small subset.
   {(Bottom) Training and validation loss across epochs (log scale) for a feedforward network with hidden layers of 64, 32, and 16 neurons trained using Levenberg–Marquardt. Both losses drop sharply in the first few epochs and then stabilize, indicating fast convergence with consistent generalization.}}
   \label{fig:nn_diagnostics}
\end{figure}

The dataset consists of 5,097 samples representing various device configurations under different magnetic field conditions. We randomly split the data into training (80\%) and testing (20\%) sets. 
{The input features used for model training include the in-plane magnetic field components $b_x$ and $b_y$, which describe the direction of the applied magnetic field, the corresponding magnetic field magnitude $|\mathbf{b}|$ representing its overall strength, the actuation frequency, and the tip distance from the coil system.}
{The target outputs correspond to the three-dimensional coordinates of the four Bezier control points ($\mathbf{p}_0$--$\mathbf{p}_3$) that parameterize the robot’s reconstructed shape from FBG measurements. These control points provide a compact geometric representation of the overall device configuration, serving as the prediction targets for both the NN and RF models.}
For the NN model, we implemented a feedforward architecture with three hidden layers (64, 32, and 16 neurons) using the Levenberg-Marquardt backpropagation algorithm. As shown in Fig.~\ref{fig:nn_diagnostics}, the model achieved optimal performance at approximately 30\% of the training data, with minimal improvement thereafter, suggesting efficient learning from a relatively small portion of the dataset. The overall correlation coefficient across all 12 output variables was $R=0.90574$. For the RF model, we trained 12 separate regressors using 200 trees per forest with a minimum leaf size of 5.
{The RF approach was selected for its robustness to measurement noise, low overfitting risk, and ability to model nonlinear input–output relationships without requiring smoothness or continuity assumptions. Each decision tree was trained on randomly sampled subsets of data and features to improve model diversity and generalization. This ensemble structure enables the RF to capture localized nonlinearities between magnetic actuation inputs and Bezier control points while maintaining computational efficiency during inference. In addition, the RF provides feature importance metrics that help identify which magnetic field components most strongly influence the robot’s deformation. Compared to the NN model, which requires careful hyperparameter tuning and larger datasets, the RF achieves high predictive accuracy with minimal tuning, making it well-suited for real-time prediction and control in surgical applications.}

\subsection{Model Performance Comparison}
\label{Model Performance Comparison}
\begin{figure}[!t]
\centering
\includegraphics[width=\linewidth]{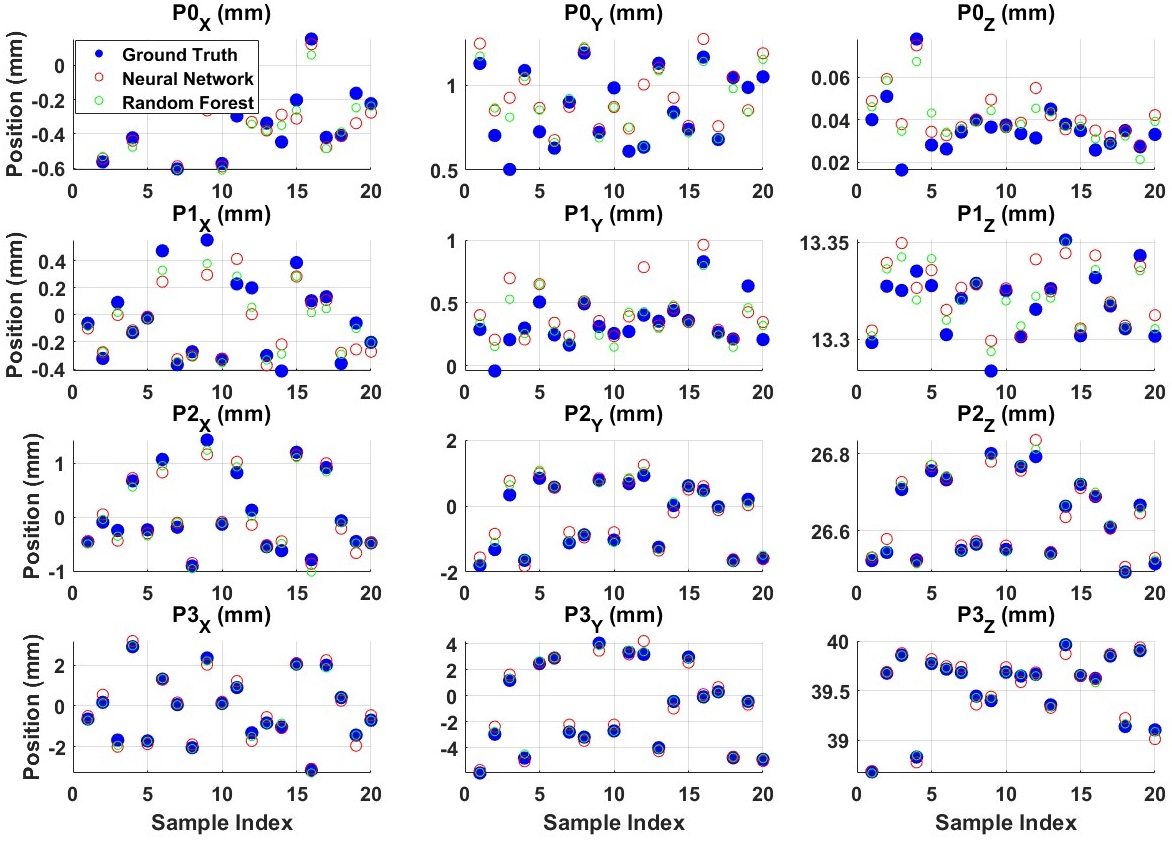}
\caption{Prediction accuracy for all Bezier control points. Blue dots: ground truth, red circles: NN predictions, green circles: RF predictions.}
\label{fig:model_prediction}
\end{figure}

\begin{figure}[!t]
\centering
\includegraphics[width=\linewidth]{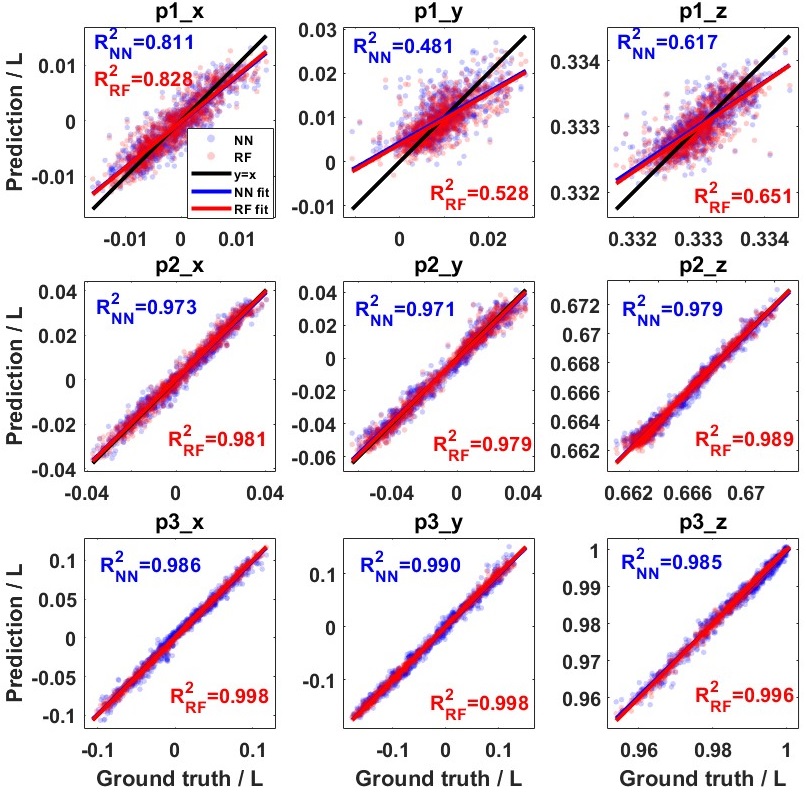}
\caption{Comparison of NN-predicted and actual Bezier control points across training, validation, testing, and full datasets.}
\label{fig:regression_analysis}
\end{figure}

\begin{figure}[!t]
\centering
\includegraphics[width=\linewidth]{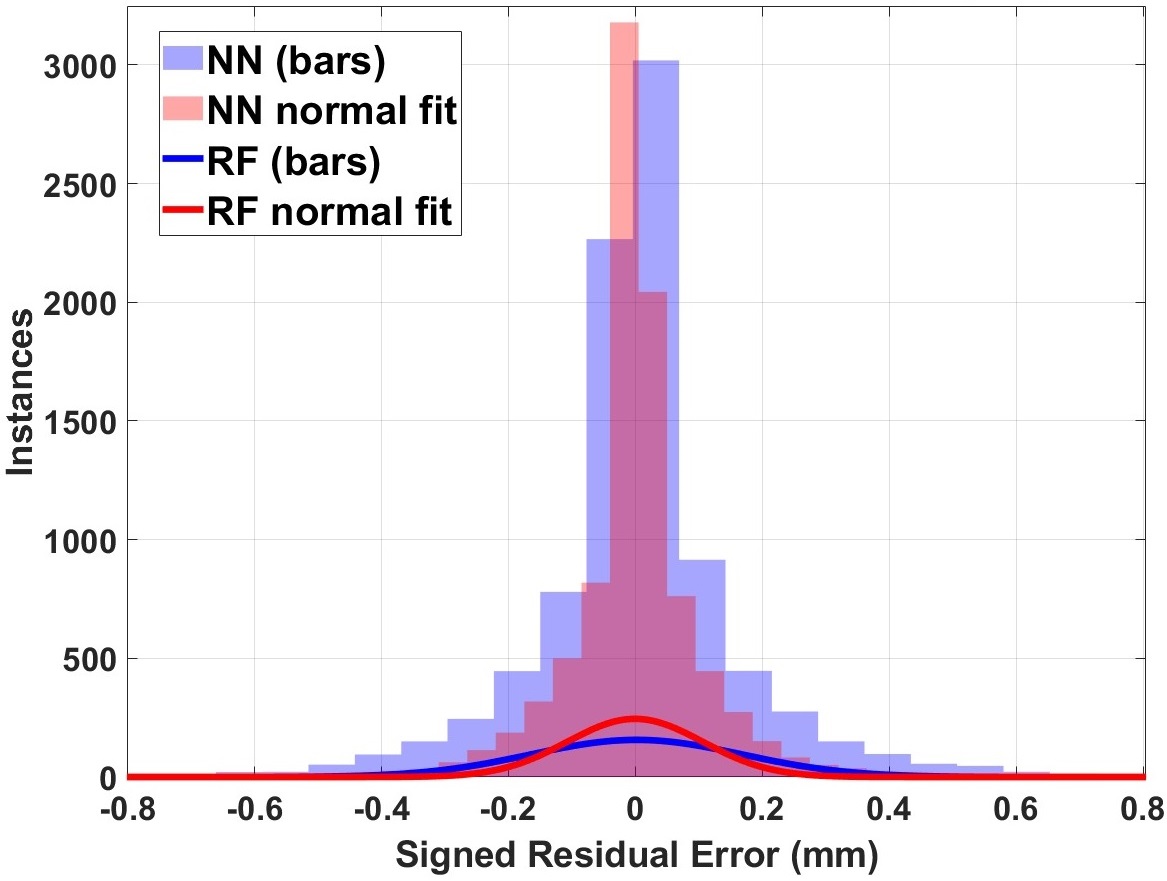}
\caption{Error histogram showing the distribution of prediction errors. The concentration of errors around zero indicates unbiased predictions.}
\label{fig:error_histogram}
\end{figure}

{Table}~\ref{tab:model_performance} {presents the detailed performance metrics of both models across all control points. As summarized in the table, the NN baseline achieved an overall RMSE of 0.121~mm (R$^2$~=~0.788, max error~=~1.56~mm), whereas the RF achieved 0.087~mm (R$^2$~=~0.816, max error~=~1.11~mm). Based on these absolute values, the RF model demonstrates a 28.37\% reduction in RMSE compared with the NN model. This improvement is particularly pronounced for the distal control points ($\mathbf{p}_2$ and $\mathbf{p}_3$), which are critical for accurate device tip positioning. The model accuracy comparison by individual control points, as shown in Figs.}~\ref{fig:model_prediction} {and} \ref{fig:regression_analysis}, {reveals that while both models predict proximal control points ($\mathbf{p}_0$ and $\mathbf{p}_1$) with similar accuracy, the RF model outperforms the NN model for distal control points ($\mathbf{p}_2$ and $\mathbf{p}_3$). Quantitatively, the RF model reduces RMSE by 5.97\% for $\mathbf{p}_0$, 5.47\% for $\mathbf{p}_1$, 16.41\% for $\mathbf{p}_2$, and an impressive 53.94\% for $\mathbf{p}_3$ compared to the NN model. The error distribution analysis further shows that both models exhibit primarily zero-centered errors (Fig.}~\ref{fig:error_histogram}){, with the RF model demonstrating tighter error distributions across all control points. The NN model struggles with $\mathbf{p}_3$ coordinates, exhibiting errors exceeding 1.5~mm for ${p}_{3Y}$, whereas the RF model maintains maximum absolute errors below 1.11~mm. These findings confirm that the RF model provides superior predictive accuracy, particularly in the distal region of the device where precise control is most critical.}

\begin{table*}[!t]
\centering
\caption{Performance comparison between NN and RF models.}
\label{tab:model_performance}
\renewcommand{\arraystretch}{1.2}
\begin{tabular*}{\textwidth}{@{\extracolsep{\fill}} c|c c c|c c c }
\hline
\multirow{2}{*}{\textbf{Points}} & \multicolumn{3}{c|}{\textbf{Neural Network}} & \multicolumn{3}{c}{\textbf{Random Forest}} \\
\cline{2-7}
 & \textbf{RMSE} & \textbf{R$^2$} & \textbf{Max Error} & \textbf{RMSE} & \textbf{R$^2$} & \textbf{Max Error} \\
 & (mm) & - & (mm) & (mm) & - & (mm) \\
\hline
$p_{0X}$ & 0.088 & 0.761 & 0.376 & 0.083 & 0.790 & 0.340 \\
$p_{0Y}$ & 0.157 & 0.428 & 0.698 & 0.148 & 0.493 & 0.743 \\
$p_{0Z}$ & 0.009 & 0.503 & 0.049 & 0.008 & 0.563 & 0.044 \\
\hline
$p_{1X}$ & 0.098 & 0.811 & 0.413 & 0.094 & 0.828 & 0.395 \\
$p_{1Y}$ & 0.167 & 0.461 & 0.803 & 0.157 & 0.524 & 0.869 \\
$p_{1Z}$ & 0.010 & 0.612 & 0.061 & 0.010 & 0.648 & 0.060 \\
\hline
$p_{2X}$ & 0.115 & 0.973 & 0.393 & 0.096 & 0.981 & 0.362 \\
$p_{2Y}$ & 0.180 & 0.971 & 0.903 & 0.152 & 0.979 & 0.917 \\
$p_{2Z}$ & 0.015 & 0.979 & 0.061 & 0.011 & 0.989 & 0.050 \\
\hline
$p_{3X}$ & 0.227 & 0.986 & 0.946 & 0.095 & 0.998 & 0.415 \\
$p_{3Y}$ & 0.336 & 0.990 & 1.562 & 0.162 & 0.998 & 1.110 \\
$p_{3Z}$ & 0.052 & 0.985 & 0.236 & 0.026 & 0.996 & 0.125 \\
\hline
\textbf{Overall} & \textbf{0.121} & \textbf{0.788} & \textbf{1.562} & \textbf{0.087} & \textbf{0.816} & \textbf{1.110} \\
\hline
\end{tabular*}
\end{table*}

\subsection{Feature Importance Analysis}
\label{Feature Importance Analysis}
To gain insights into the relationship between magnetic field parameters and device deformation, we analyzed the feature importance metrics derived from the RF model. Our results revealed distinct patterns of influence across different control points: The magnetic field components ($b_x$ and $b_y$) dominate the influence on most control point coordinates, with $b_y$ showing particularly high importance ($>$50\%) for ${p}_{0Y}$, ${p}_{2Y}$, and ${p}_{3Y}$ coordinates. This aligns with the physical understanding of magnetic actuation, where the Y-component of the magnetic field primarily controls lateral movement of the device. 
{The frequency parameter exhibits uniquely high importance (32.34\%) for ${p}_{0Z}$, suggesting that actuation frequency influences the vertical position (z-coordinate) of the device’s proximal segment, i.e., the slight vertical displacement at the base resulting from dynamic bending under oscillating magnetic fields.}
The distance parameter shows consistent influence on $\mathbf{p}_0$ coordinates (23.81\% for ${p}_{0X}$, 24.50\% for ${p}_{0Z}$) and $\mathbf{p}_3$ coordinates (13.00\% for ${p}_{3X}$, 13.17\% for ${p}_{3Y}$, 15.79\% for ${p}_{3Z}$), reflecting how the proximity to the electromagnetic coil system affects both the fixed base and the free tip of the device. This feature importance analysis provides valuable insights for control system design, indicating which magnetic field parameters should be prioritized for precise manipulation of specific device coordinates. For instance, fine control of the device tip position would benefit from careful modulation of both $b_x$ and $b_y$, while actuation frequency could be leveraged for height adjustment at the device base.

\subsection{Bezier Curve Shape Reconstruction}
\label{Bezier Curve Shape Reconstruction}
\begin{figure}[!t]
  \centering
  \includegraphics[width=\linewidth]{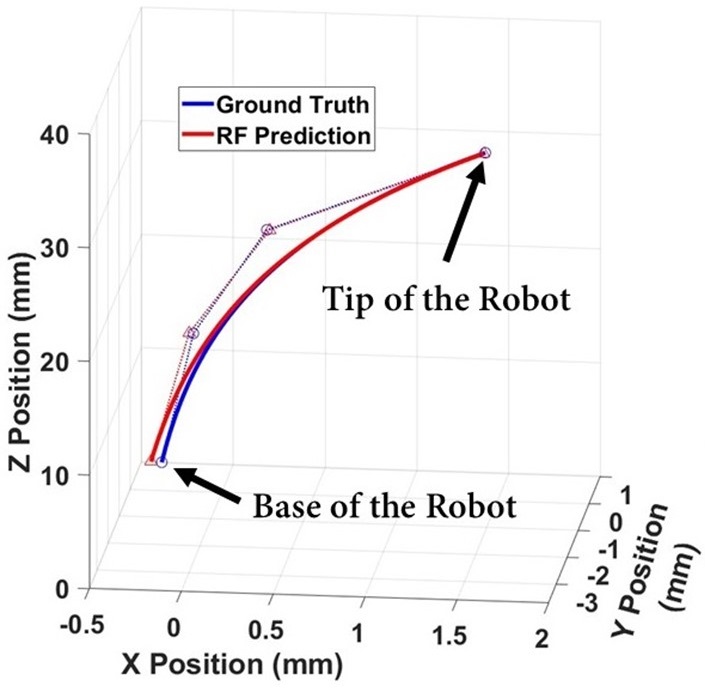}
  \caption{{Bezier curve shape prediction (3D view). Ground truth shown in blue and RF predictions shown in red. The close alignment between the predicted and measured curves demonstrates the RF model’s strong capability to capture complex nonlinear deformations of the soft suction device. The figure highlights the base and tip of the robot, illustrating how the predicted Bezier curve accurately preserves both overall curvature and tip position within sub-millimeter deviation.}}
  \label{fig:shape_prediction_3d}
\end{figure}

\begin{figure}[!t]
  \centering
  \includegraphics[width=0.8\linewidth]{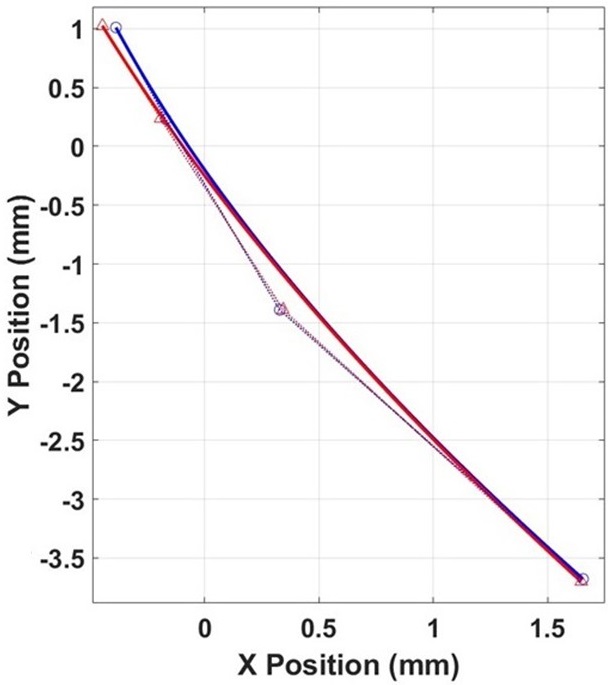}
  \caption{Bezier curve shape prediction (side view). Ground truth (blue) vs.\ RF predictions (green).}
  \label{fig:shape_prediction_side}
  \vspace{-2 mm}
\end{figure}

Beyond individual control point accuracy, we evaluated the overall shape reconstruction performance by comparing the Bezier curves generated from predicted control points with ground truth curves. Figures \ref{fig:shape_prediction_3d} and \ref{fig:shape_prediction_side} illustrate this comparison from both 3D and top-view perspectives. The 3D visualization confirms that the RF model achieves excellent shape reconstruction across diverse deformation patterns. Predicted Bezier curves follow the ground-truth trajectories closely, preserving control-point placement and smooth interpolation. Quantitatively, the RF delivers a MAE of $0.064\;\text{mm}$ and a $95^{\text{th}}$-percentile error of $0.225\;\text{mm}$, both negligible relative to the device diameter (4 mm) and therefore suitable for safe navigation around delicate skull-base anatomy during endoscopic endonasal procedures.

\begin{table}[!t]
\centering
\caption{Comparison of learning models on the test set.}
\label{tab:model_summary}
\small
\setlength{\tabcolsep}{3.5pt}
\renewcommand{\arraystretch}{1.2}
\begin{tabular}{c|cccc|cc}
\hline
\multirow{2}{*}{\textbf{Model}} & \multicolumn{4}{c|}{\textbf{Error metrics}} & \multicolumn{2}{c}{\textbf{Timing}} \\
\cline{2-7}
 & \textbf{RMSE} & $\mathbf{R^{2}}$ & \textbf{Mean} & \textbf{SD} & \textbf{Train} & \textbf{Pred /} \\
 & \textbf{(mm)} &  & \textbf{$\sum e^{2}$} & \textbf{$\sum e^{2}$} & \textbf{(s)} & \textbf{sample (s)} \\
\hline
NN & 0.121  & 0.788 & 0.283 & 0.329 & 176.2 & $1.07\times10^{-4}$ \\
RF & 0.0868 & 0.816 & 0.131 & 0.204 &  26.9 & $7.89\times10^{-5}$ \\
\hline
\end{tabular}
\end{table}

The regression analysis of the NN model (Fig.~\ref{fig:regression_analysis}) yielded correlation coefficients of $R_\text{train}=0.9169$, $R_\text{val}=0.8815$, and $R_\text{test}=0.8782$, indicating good generalization with limited over-fitting. Nevertheless, the RF model surpassed the NN across every metric. Using the test set, we formed per-sample total errors $\sum_{i=1}^{12}(y_i-\hat y_i)^2$ for both models.  Lilliefors tests detected slight departures from perfect normality for NN and RF residuals ($p=0.001$ for each). Levene’s test showed unequal variances ($p=4.96\times10^{-21}$); consequently, Welch’s two-sample $t$-test was applied, revealing a highly significant reduction in error for RF compared with NN ($t=12.55$, $p=1.32\times10^{-34}$). Mean ($\pm$\,SD) total error fell from $0.283\pm0.329$ for the NN to $0.131\pm0.204$ for the RF (Table~\ref{tab:model_summary}). These findings confirm that model selection materially influences residual error and reinforce our choice of RF for subsequent real-time deployment.

\subsection{FBG Ground Truth Limitations and Accuracy}
\label{FBG Ground Truth Limitations and Accuracy}
The accuracy validation methodology of this study warrants careful examination, particularly regarding the use of reconstructed shapes from FBG as ground truth for training the learning based models. While multi-core fiber Bragg grating sensors are well-established for shape sensing in medical robotics applications, their inherent limitations must be acknowledged when interpreting the reported accuracy metrics. 
{The FBG system employed in this study uses a multi-core fiber containing 26 gratings uniformly spaced at 10~mm intervals along a total sensing length of 250~mm, with approximately five gratings located within the 40~mm segment embedded in the soft body of the device. This configuration provides sufficient spatial resolution for curvature reconstruction. {To provide context regarding sensor resolution capabilities in FBG-based shape sensing, we note that Dong et~al.}~\cite{dong2022shape} {employed a denser grating configuration (3.17~mm spacing) and reported 1.53~mm tip position accuracy when validated against electromagnetic tracking. While their experimental setup, validation methodology, and application context differ from the present study, this reference illustrates that finer grating resolution combined with independent external validation can achieve higher absolute positioning accuracy. The sensing layout used here is comparatively coarser, which motivated our decision to report model-to-FBG consistency rather than absolute shape accuracy.}}
The piecewise constant curvature assumption between grating points introduces interpolation errors that accumulate along the fiber length, potentially resulting in tip position uncertainties exceeding 1~mm despite the sensor's high strain sensitivity.

{The reported sub-millimeter accuracy (0.064 mm RMSE) quantifies the fidelity of the Bezier curve representation relative to the FBG-reconstructed shape, confirming that the four-point parameterization accurately captures the device’s geometry. The RF model’s prediction accuracy for these control points, reported in Table}~\ref{tab:model_performance}, {achieved an overall RMSE of 0.087 mm. The RMSE used throughout this study was computed as:}
\begin{equation}
\mathrm{RMSE} = \sqrt{\frac{1}{N} \sum_{i=1}^{N} \left\| \mathbf{y}_i - \hat{\mathbf{y}}_i \right\|^2}
\label{eq:rmse}
\end{equation}
where $\mathbf{y}_i$ and $\hat{\mathbf{y}}_i$ denote the ground-truth and predicted 3D point coordinates, respectively, and $N$ is the total number of evaluated points. 
This metric should not be interpreted as absolute shape sensing accuracy, but rather as the model's capacity to learn and reproduce the mapping between magnetic field inputs and FBG-measured deformations. The distinction is critical: while the learning-based approach successfully captures the nonlinear relationship between actuation and measured response with high fidelity, the absolute accuracy of the predicted shapes relative to true physical deformation remains bounded by the FBG system's inherent limitations. {It should be noted that FBG shape reconstruction employs continuous shape reconstruction, implemented via piecewise constant curvature (PCC) interpolation between discrete grating locations, which is standard practice in FBG-based sensing systems. This is fundamentally different from traditional PCC-based kinematic modeling, where constant-curvature assumptions are used to analytically derive actuation-to-shape relationships. Our learning framework captures experimentally observed behaviors, including frequency-dependent and viscoelastic effects, that analytical PCC kinematic models cannot represent.}

{This distinction is further evidenced by our experimental results, where the same magnetic field magnitude produces different deformation patterns at different actuation frequencies (as shown in Fig.}~\ref{fig:xy_trajs}{), a dynamic behavior that the learning model successfully captures but that analytical PCC kinematic models fundamentally cannot represent. The learning model thus encodes richer behavioral information than what analytical PCC-based models could provide, even though both share a common continuous shape reconstruction step.}

{Future work will incorporate independent validation using calibrated vision systems or higher-resolution tracking methods to establish absolute accuracy bounds. Despite these limitations, the learning-based approach remains valid for the intended application of real-time shape estimation, where consistent relative deformation tracking is more critical than absolute position accuracy. The model's ability to predict deformation patterns with high repeatability demonstrates strong potential for integration into future closed-loop control frameworks, where feedback from embedded FBG sensing could be leveraged to compensate for systematic errors and improve precision during surgical manipulation. {Additionally, future work could explore hybrid modeling strategies that combine the theoretical rigor of physics-based frameworks such as the GVS approach}~\cite{mathew2025reduced} {with data-driven components for parameter identification or model correction, potentially offering improved generalizability while retaining the empirical accuracy demonstrated in this study.}}

Furthermore, this study utilized only a single physical prototype with data collected across 70 actuation scenarios, preventing assessment of device-to-device variability or manufacturing tolerances. The learning-based model is inherently device-specific, trained on this particular prototype's characteristics without validation across multiple devices. Clinical translation would necessitate either individual calibration for each manufactured device or training across multiple prototypes to establish model robustness. While the current work demonstrates the feasibility of the learning-based approach for shape prediction, comprehensive validation across multiple devices with varying material properties and manufacturing variations remains essential for establishing clinical applicability and generalization capabilities of the proposed methodology.

{Also, the rotating magnetic field profiles employed in this study were selected to provide controlled and repeatable deformation trajectories while sampling different magnetic field directions across the coil workspace. As the tip moves through circular paths, the local magnetic field vectors vary spatially around the coil table, enabling the model to learn deformation responses under multiple actuation orientations. Additionally, the experiments were conducted across a frequency range of 0.2–1.0~Hz to capture the dynamic, viscoelastic behavior of the soft body under time-varying loads. This combined spatial and temporal variation provided a rich dataset for model training and validation. Future work will extend this framework to include non-circular and task-oriented magnetic actuation profiles, such as targeted tip positioning and complex navigation trajectories, to further evaluate the model’s generalization in conditions that closely resemble surgical operations.}

Lastly, the current study trained and evaluated the model exclusively using uniform, symmetric rotating magnetic fields generated under controlled laboratory conditions. Clinical environments present significantly more complex magnetic field profiles, including spatial non-uniformities, field gradients, and asymmetric distributions caused by nearby ferromagnetic materials, imaging equipment, or electromagnetic interference. The learning-based model's performance under such irregular field conditions remains untested and would likely degrade without prior exposure to these variations during training. Robust clinical deployment would require expanding the training dataset to encompass diverse magnetic field profiles representative of surgical environments, including both measured clinical field distributions and systematically varied asymmetric patterns. This expanded training approach would enable the model to learn invariant features across field variations rather than overfitting to idealized uniform fields.

\subsection{{Generalization to confined surgical environments}}
{The experiments presented in this work were performed in an open-space setting to isolate and characterize the fundamental deformation behavior of the magnetically actuated soft suction device. This environment provided a controlled platform for capturing the intrinsic relationship between magnetic actuation inputs and device deformation without external constraints, ensuring consistent data for model training and validation. However, in realistic endoscopic endonasal surgery, the device will operate within narrow anatomical corridors surrounded by rigid and soft tissues. These boundaries can restrict bending, cause asymmetric deformation, and introduce frictional or damping effects due to tissue contact and mucus layers. We acknowledge that such confinement and viscous interaction could alter the deformation patterns observed under open-space conditions. Therefore, the current results represent a baseline characterization of the device’s intrinsic magnetic response, which serves as a necessary first step before introducing the additional complexities of tissue interaction. To extend the modeling framework toward clinically relevant environments, future work will incorporate training data acquired within anatomically realistic phantoms that replicate skull-base geometries and the mechanical properties of nasal tissues. The existing Bezier-based learning model is compatible with such data expansion because it encodes shape using a compact set of control points that can adapt to new boundary conditions. During these experiments, additional input features such as contact location, frictional interaction indicators, or limited endoscopic vision feedback may be included to improve model robustness under partial confinement. Furthermore, the integration of real-time FBG sensing with image-based feedback could enable adaptive online correction of the predicted shape during surgical manipulation. Through these planned steps, the proposed learning-based framework can be systematically generalized to accurately predict and control device behavior in the constrained and viscoelastic environments encountered in actual endonasal neurosurgery.}

\section{Conclusion}
\label{Conclusion}
This paper presented {a generalized learning-based modeling framework that transforms FBG shape sensing data into Bezier control points, establishing a new abstraction for modeling magnetically actuated soft surgical instruments.} {This framework introduces a unified methodology that links magnetic actuation inputs, embedded sensing, and Bezier-based geometric representation into an interpretable and compact modeling structure.} This approach bridges the gap between shape sensing technologies and practical control applications by providing a direct mapping between the magnetic field and the resulting device configuration. Our key contribution was the development and validation of a framework that enables highly accurate real-time shape prediction of complex non-linear deformations in soft robotic devices. By comparing NN and RF modeling approaches, we demonstrated that ensemble-based methods outperform traditional NN for this application, particularly at the critical distal end of the device, where precise control is essential for surgical tasks. Also, the computational efficiency of our approach supports high-frequency real-time control suitable for dynamic surgical environments, while maintaining sub-millimeter accuracy across the device's entire structure. {This study establishes a new modeling paradigm that integrates magnetic actuation, data-driven learning, and geometric Bezier parameterization, forming a transferable foundation for future control and design of magnetically actuated continuum robots.}
Future work will focus on three main directions: (1) integrating this modeling approach into a closed-loop control system for autonomous navigation, (2) expanding the training dataset to include additional device configurations for enhanced generalization, {including exploration of hybrid physics-informed learning approaches that leverage established frameworks such as GVS for improved interpretability and generalizability,} and (3) validating the model through ex vivo experiments in simulated surgical scenarios.

\section*{Funding Source Statement}
This work was supported by the Canadian Institutes of Health Research (CIHR) Project Grant 452287.

\section*{Declaration of Competing Interest}
The authors declare that they have no competing financial interests or personal relationships that could have influenced the work reported in this paper.

\section*{Declaration of Generative AI Use}
The author(s) used ChatGPT for grammar and language checking only. No scientific content was generated using AI. The author(s) reviewed and edited the content and take full responsibility for the published article.

\section*{Data Availability}
Data will be made available on request.

\bibliographystyle{elsarticle-num}
\bibliography{Main.bib}

@inproceedings{deng2024towards,
  title={Towards Bimanual Operation of Magnetically Actuated Surgical Instruments},
  author={Deng, Yuanzhe and Roshanfar, Majid and Mayer, Haley and He, Changyan and Drake, James and Looi, Thomas and Diller, Eric},
  booktitle={2024 10th IEEE RAS/EMBS International Conference for Biomedical Robotics and Biomechatronics (BioRob)},
  pages={1295--1300},
  year={2024},
  organization={IEEE}
}

@article{sasagawa2024endoscopic,
  title={Endoscopic and exoscopic surgery for brain tumors},
  author={Sasagawa, Yasuo and Tanaka, Shingo and Kinoshita, Masashi and Nakada, Mitsutoshi},
  journal={International Journal of Clinical Oncology},
  volume={29},
  number={10},
  pages={1399--1406},
  year={2024},
  publisher={Springer}
}

@article{valencia2024special,
  title={Special Considerations in Pediatric Endoscopic Skull Base Surgery},
  author={Valencia-Sanchez, Bastien A and Kim, Jeeho D and Zhou, Sheng and Chen, Sonja and Levy, Michael L and Roxbury, Christopher and Patel, Vijay A and Polster, Sean P},
  journal={Journal of Clinical Medicine},
  volume={13},
  number={7},
  pages={1924},
  year={2024},
  publisher={MDPI}
}

@article{yang2023magnetically,
  title={Magnetically actuated continuum medical robots: A review},
  author={Yang, Zhengxin and Yang, Haojin and Cao, Yanfei and Cui, Yaoyao and Zhang, Li},
  journal={Advanced intelligent systems},
  volume={5},
  number={6},
  pages={2200416},
  year={2023},
  publisher={Wiley Online Library}
}

@inproceedings{roshanfar2025soft,
  title={Soft Magnetically Steerable Suction Device for Endoscopic Endonasal Brain Tumor Extraction},
  author={Roshanfar, Majid and Zhang, Alex and He, Changyan and Hooshiar, Amir and Drake, James and Looi, Thomas and Diller, Eric},
  booktitle={2025 IEEE 8th International Conference on Soft Robotics (RoboSoft)},
  pages={1--6},
  year={2025},
  organization={IEEE}
}

@article{schonewille2024electromagnets,
  title={Electromagnets under the table: an unobtrusive magnetic navigation system for microsurgery},
  author={Schonewille, Adam and He, Changyan and Forbrigger, Cameron and Wu, Nancy and Drake, James and Looi, Thomas and Diller, Eric},
  journal={IEEE Transactions on Medical Robotics and Bionics},
  volume={6},
  number={3},
  pages={980--991},
  year={2024},
  publisher={IEEE}
}

@article{roshanfar2025advanced,
  title={Advanced Robotics for the Next-Generation of Cardiac Interventions},
  author={Roshanfar, Majid and Salimi, Mohammadhossein and Kaboodrangi, Amir Hossein and Jang, Sun-Joo and Sinusas, Albert J and Wong, Shing-Chiu and Mosadegh, Bobak},
  journal={Micromachines},
  volume={16},
  number={4},
  pages={363},
  year={2025},
  publisher={MDPI}
}

@article{dreyfus2024dexterous,
  title={Dexterous helical magnetic robot for improved endovascular access},
  author={Dreyfus, Roland and Boehler, Quentin and Lyttle, Sean and Gruber, Philipp and Lussi, Jonas and Chautems, Christophe and Gervasoni, Simone and Berberat, Jatta and Seibold, Dominic and Ochsenbein-K{\"o}lble, Nicole and others},
  journal={Science Robotics},
  volume={9},
  number={87},
  pages={eadh0298},
  year={2024},
  publisher={American Association for the Advancement of Science}
}

@article{kim2019endoscopic,
  title={Endoscopic endonasal skull base surgery for pediatric brain tumors},
  author={Kim, Yong Hwy and Lee, Ji Yeoun and Phi, Ji Hoon and Wang, Kyu-Chang and Kim, Seung-Ki},
  journal={Child's Nervous System},
  volume={35},
  pages={2081--2090},
  year={2019},
  publisher={Springer}
}

@article{chen2022magnetically,
  title={Magnetically actuated capsule robots: A review},
  author={Chen, Weiyuan and Sui, Jianbo and Wang, Chengyong},
  journal={IEEE Access},
  volume={10},
  pages={88398--88420},
  year={2022},
  publisher={IEEE}
}

@inproceedings{hooshiar2020accurate,
  title={Accurate estimation of tip force on tendon-driven catheters using inverse cosserat rod model},
  author={Hooshiar, Amir and Sayadi, Amir and Jolaei, Mohammad and Dargahi, Javad},
  booktitle={2020 International Conference on Biomedical Innovations and Applications (BIA)},
  pages={37--40},
  year={2020},
  organization={IEEE}
}

@inproceedings{li2023dexterity,
  title={Dexterity of Concentric Magnetic Continuum Robot with Multiple Stiffness},
  author={Li, Na and Lin, Daojing and Wu, Junfeng and Gan, Quan and Jiao, Niandong},
  booktitle={International Conference on Intelligent Robotics and Applications},
  pages={329--338},
  year={2023},
  organization={Springer}
}

@article{hong2020magnetic,
  title={Magnetic control of a flexible needle in neurosurgery},
  author={Hong, Ayoung and Petruska, Andrew J and Zemmar, Ajmal and Nelson, Bradley J},
  journal={IEEE Transactions on Biomedical Engineering},
  volume={68},
  number={2},
  pages={616--627},
  year={2020},
  publisher={IEEE}
}

@article{test,
  title={Ferromagnetic soft continuum robots},
  author={Kim, Yoonho and Parada, German A and Liu, Shengduo and Zhao, Xuanhe},
  journal={Science Robotics},
  volume={4},
  number={33},
  pages={eaax7329},
  year={2019},
  publisher={American Association for the Advancement of Science}
}

@article{he2025magnetically,
  title={Magnetically actuated dexterous tools for minimally invasive operation inside the brain},
  author={He, Changyan and Nguyen, Robert and Mayer, Haley and Cheng, Lingbo and Kang, Paul and Aubeeluck, D Anastasia and Thiong’o, Grace and Fredin, Erik and Drake, James and Looi, Thomas and others},
  journal={Science Robotics},
  volume={10},
  number={100},
  pages={eadk4249},
  year={2025},
  publisher={American Association for the Advancement of Science}
}

@article{abbott2020magnetic,
  title={Magnetic methods in robotics},
  author={Abbott, Jake J and Diller, Eric and Petruska, Andrew J},
  journal={Annual Review of Control, Robotics, and Autonomous Systems},
  volume={3},
  number={1},
  pages={57--90},
  year={2020},
  publisher={Annual Reviews}
}

@article{Zhao_Ferromag_SciRob19,
  title={Ferromagnetic soft continuum robots},
  author={Kim, Yoonho and Parada, German A and Liu, Shengduo and Zhao, Xuanhe},
  journal={Science Robotics},
  volume={4},
  number={33},
  pages={eaax7329},
  year={2019},
  publisher={American Association for the Advancement of Science}
}

@ARTICLE{Nelson_CardiacCatheter18,
  author={Chautems, Christophe and Lyttle, Sean and Boehler, Quentin and Nelson, Bradley J.},
  journal={IEEE Robotics and Automation Letters}, 
  title={Design and Evaluation of a Steerable Magnetic Sheath for Cardiac Ablations}, 
  year={2018},
  volume={3},
  number={3},
  pages={2123-2128},
  doi={10.1109/LRA.2018.2809546}}

@article{rox2020mechatronic,
  title={Mechatronic design of a two-arm concentric tube robot system for rigid neuroendoscopy},
  author={Rox, Margaret F and Ropella, Dominick S and Hendrick, Richard J and Blum, Evan and Naftel, Robert P and Bow, Hansen C and Herrell, S Duke and Weaver, Kyle D and Chambless, Lola B and Webster III, Robert J},
  journal={IEEE/ASME transactions on mechatronics},
  volume={25},
  number={3},
  pages={1432--1443},
  year={2020},
  publisher={IEEE}
}

@article{masoumi2024embedded,
  title={Embedded Force Sensor for Soft Robots With Deep Transformation Calibration},
  author={Masoumi, Navid and Ramos, Andr{\'e}s C and Torkaman, Tannaz and Feldman, Liane S and Barralet, Jake and Dargahi, Javad and Hooshiar, Amir},
  journal={IEEE Transactions on Medical Robotics and Bionics},
  year={2024},
  publisher={IEEE}
}

@article{torkaman2023embedded,
  title={Embedded six-dof force--torque sensor for soft robots with learning-based calibration},
  author={Torkaman, Tannaz and Roshanfar, Majid and Dargahi, Javad and Hooshiar, Amir},
  journal={IEEE Sensors Journal},
  volume={23},
  number={4},
  pages={4204--4215},
  year={2023},
  publisher={IEEE}
}

@inproceedings{jolaei2020sensor,
  title={Sensor-free force control of tendon-driven ablation catheters through position control and contact modeling},
  author={Jolaei, Mohammad and Hooshiar, Amir and Sayadi, Amir and Dargahi, Javad and Packirisamy, Muthukumaran},
  booktitle={2020 42nd annual international conference of the ieee engineering in medicine \& biology society (EMBC)},
  pages={5248--5251},
  year={2020},
  organization={IEEE}
}

@article{roshanfar2023hyperelastic,
  title={Hyperelastic modeling and validation of hybrid-actuated soft robot with pressure-stiffening},
  author={Roshanfar, Majid and Taki, Salar and Sayadi, Amir and Cecere, Renzo and Dargahi, Javad and Hooshiar, Amir},
  journal={Micromachines},
  volume={14},
  number={5},
  pages={900},
  year={2023},
  publisher={MDPI}
}

@article{al2021fbg,
  title={FBG-based estimation of external forces along flexible instrument bodies},
  author={Al-Ahmad, Omar and Ourak, Mouloud and Vlekken, Johan and Vander Poorten, Emmanuel},
  journal={Frontiers in Robotics and AI},
  volume={8},
  pages={718033},
  year={2021},
  publisher={Frontiers Media SA}
}

@article{pittiglio2022patient,
  title={Patient-specific magnetic catheters for atraumatic autonomous endoscopy},
  author={Pittiglio, Giovanni and Lloyd, Peter and da Veiga, Tomas and Onaizah, Onaizah and Pompili, Cecilia and Chandler, James H and Valdastri, Pietro},
  journal={Soft robotics},
  volume={9},
  number={6},
  pages={1120--1133},
  year={2022},
  publisher={Mary Ann Liebert, Inc., publishers 140 Huguenot Street, 3rd Floor New~…}
}

@article{burgner2013telerobotic,
  title={A telerobotic system for transnasal surgery},
  author={Burgner, Jessica and Rucker, D Caleb and Gilbert, Hunter B and Swaney, Philip J and Russell, Paul T and Weaver, Kyle D and Webster, Robert J},
  journal={IEEE/ASME transactions on mechatronics},
  volume={19},
  number={3},
  pages={996--1006},
  year={2013},
  publisher={IEEE}
}

@article{martinez2021modern,
  title={Modern endoscopic skull base neurosurgery},
  author={Martinez-Perez, Rafael and Requena, Luis C and Carrau, Ricardo L and Prevedello, Daniel M},
  journal={Journal of neuro-oncology},
  volume={151},
  pages={461--475},
  year={2021},
  publisher={Springer}
}

@misc{carbon_sil30_tds,
  author       = {{Carbon, Inc.}},
  title        = {{SIL 30 Technical Datasheet}},
  year         = {2024},
  note         = {Doc \#106453-00 Rev F},
  url          = {https://www.carbon3d.com/materials/sil-30}
}

@article{sayadi2024finite,
  title={Finite arc method: fast-solution extended piecewise constant curvature model of soft robots with large variable curvature deformations},
  author={Sayadi, Amir and Cecere, Renzo and Hooshiar, Amir},
  journal={Robotics Reports},
  volume={2},
  number={1},
  pages={49--64},
  year={2024},
  publisher={Mary Ann Liebert, Inc., publishers 140 Huguenot Street, 3rd Floor New~…}
}

@article{gholami2025advances,
  title={Advances and Challenges in Endoscopic Endonasal Tumor Surgery: Lessons Learned from Modern Practice},
  author={Gholami, Poorya and Rezvanfar, Kiana and Valizadehashenaabad, Shahrzad and Esmaeili, Elham and Rafizadeh, Fatemeh and Aminiyan, Armin and Fakhr, Masoud Saadat},
  journal={Indian Journal of Otolaryngology and Head \& Neck Surgery},
  pages={1--22},
  year={2025},
  publisher={Springer}
}

@inproceedings{roshanfar2022stiffness,
  title={Stiffness adaptation of a hybrid soft surgical robot for improved safety in interventional surgery},
  author={Roshanfar, Majid and Sayadi, Amir and Dargahi, Javad and Hooshiar, Amir},
  booktitle={2022 44th Annual International Conference of the IEEE Engineering in Medicine \& Biology Society (EMBC)},
  pages={4853--4859},
  year={2022},
  organization={IEEE}
}

@inproceedings{roshanfar20253,
  title={3-Dof Magnetically Actuated Robotic Manipulator for Precision Soft Tissue Resection},
  author={Roshanfar, Majid and He, Changyan and Huang, Long and Li, Zhaoxin and Cheng, Lingbo and Podolsky, Dale J and Looi, Thomas and Diller, Eric},
  booktitle={2025 International Conference on Manipulation, Automation and Robotics at Small Scales (MARSS)},
  pages={1--6},
  year={2025},
  organization={IEEE}
}

@inproceedings{lee2025magnetic,
  title={Magnetic Capsule for Stable Collection of Large GI Tract Microbiome Samples},
  author={Lee, Taeyoung and Diller, Eric},
  booktitle={2025 International Conference on Manipulation, Automation and Robotics at Small Scales (MARSS)},
  pages={1--7},
  year={2025},
  organization={IEEE}
}

@article{dong2022shape,
  title={Shape tracking and feedback control of cardiac catheter using MRI-guided robotic platform—validation with pulmonary vein isolation simulator in MRI},
  author={Dong, Ziyang and Wang, Xiaomei and Fang, Ge and He, Zhuoliang and Ho, Justin Di-Lang and Cheung, Chim-Lee and Tang, Wai Lun and Xie, Xiaochen and Liang, Liyuan and Chang, Hing-Chiu and others},
  journal={IEEE Transactions on Robotics},
  volume={38},
  number={5},
  pages={2781--2798},
  year={2022},
  publisher={IEEE}
}

@article{falotico2025learning,
  title={Learning controllers for continuum soft manipulators: Impact of modeling and looming challenges},
  author={Falotico, Egidio and Donato, Enrico and Alessi, Carlo and Setti, Elisa and Nazeer, Muhammad Sunny and Agabiti, Camilla and Caradonna, Daniele and Bianchi, Diego and Piqu{\'e}, Francesco and Ansari, Yasmin Tauqeer and others},
  journal={Advanced Intelligent Systems},
  volume={7},
  number={2},
  pages={2400344},
  year={2025},
  publisher={Wiley Online Library}
}

@article{wang2021survey,
  title={A survey for machine learning-based control of continuum robots},
  author={Wang, Xiaomei and Li, Yingqi and Kwok, Ka-Wai},
  journal={Frontiers in Robotics and AI},
  volume={8},
  pages={730330},
  year={2021},
  publisher={Frontiers Media SA}
}

@article{webster2010design,
  title={Design and kinematic modeling of constant curvature continuum robots: A review},
  author={Webster III, Robert J and Jones, Bryan A},
  journal={The International Journal of Robotics Research},
  volume={29},
  number={13},
  pages={1661--1683},
  year={2010},
  publisher={SAGE Publications Sage UK: London, England}
}

@article{rucker2011statics,
  title={Statics and dynamics of continuum robots with general tendon routing and external loading},
  author={Rucker, D Caleb and Webster III, Robert J},
  journal={IEEE Transactions on Robotics},
  volume={27},
  number={6},
  pages={1033--1044},
  year={2011},
  publisher={IEEE}
}

@article{till2019real,
  title={Real-time dynamics of soft and continuum robots based on Cosserat rod models},
  author={Till, John and Aloi, Vincent and Rucker, Caleb},
  journal={The International Journal of Robotics Research},
  volume={38},
  number={6},
  pages={723--746},
  year={2019},
  publisher={SAGE Publications Sage UK: London, England}
}

@article{mathew2025reduced,
  title={Reduced order modeling of hybrid soft-rigid robots using global, local, and state-dependent strain parameterization},
  author={Mathew, Anup Teejo and Feliu-Talegon, Daniel and Alkayas, Abdulaziz Y and Boyer, Frederic and Renda, Federico},
  journal={The International Journal of Robotics Research},
  volume={44},
  number={1},
  pages={129--154},
  year={2025},
  publisher={SAGE Publications Sage UK: London, England}
}

@article{godage2015modal,
  title={Modal kinematics for multisection continuum arms},
  author={Godage, Isuru S and Medrano-Cerda, Gustavo A and Branson, David T and Guglielmino, Emanuele and Caldwell, Darwin G},
  journal={Bioinspiration \& Biomimetics},
  volume={10},
  number={3},
  pages={035002},
  year={2015},
  publisher={IOP Publishing}
}

@article{dellasantina2023soft,
  title={Model-based control of soft robots: A survey of the state of the art and open challenges},
  author={Della Santina, Cosimo and Duriez, Christian and Rus, Daniela},
  journal={IEEE Control Systems Magazine},
  volume={43},
  number={3},
  pages={30--65},
  year={2023},
  publisher={IEEE}
}

@article{almanzor2023static,
  title={Static shape control of soft continuum robots using deep visual inverse kinematic models},
  author={Almanzor, Elijah and Ye, Fan and Shi, Jialei and Thuruthel, Thomas George and Wurdemann, Helge A and Iida, Fumiya},
  journal={IEEE Transactions on Robotics},
  volume={39},
  number={4},
  pages={2973--2988},
  year={2023},
  publisher={IEEE}
}

@article{georgethuruthel2024visuo,
  title={Visuo-dynamic self-modelling of soft robotic systems},
  author={Marques Monteiro, Richard and Shi, Jialei and Wurdemann, Helge and Iida, Fumiya and George Thuruthel, Thomas},
  journal={Frontiers in Robotics and AI},
  volume={11},
  pages={1403733},
  year={2024},
  publisher={Frontiers}
}

\newpage
\onecolumn

\noindent
\begin{minipage}[t]{0.12\textwidth}
\vspace{0pt}
\includegraphics[width=0.9in,height=1in]{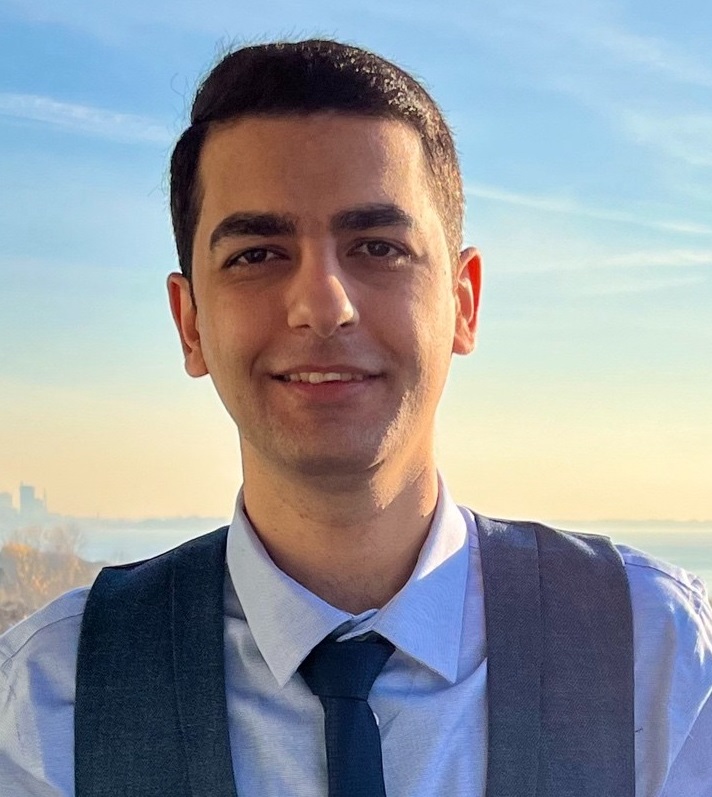}
\end{minipage}\hfill
\begin{minipage}[t]{0.85\textwidth}
\vspace{0pt}
\textbf{Majid Roshanfar} is a Postdoctoral Research Fellow at The Hospital for Sick Children (SickKids) in Toronto, Canada, and the University of Toronto. He received his Ph.D. in Mechanical Engineering from Concordia University, where his research focused on hybrid-actuated soft robots with stiffness adaptation for surgical applications. His work lies at the intersection of surgical robotics, soft and continuum robots, magnetic actuation, and shape/force sensing for minimally invasive interventions.
\end{minipage}

\vspace{10 pt}

\noindent
\begin{minipage}[t]{0.12\textwidth}
\vspace{0pt}
\includegraphics[width=0.8in,height=1in]{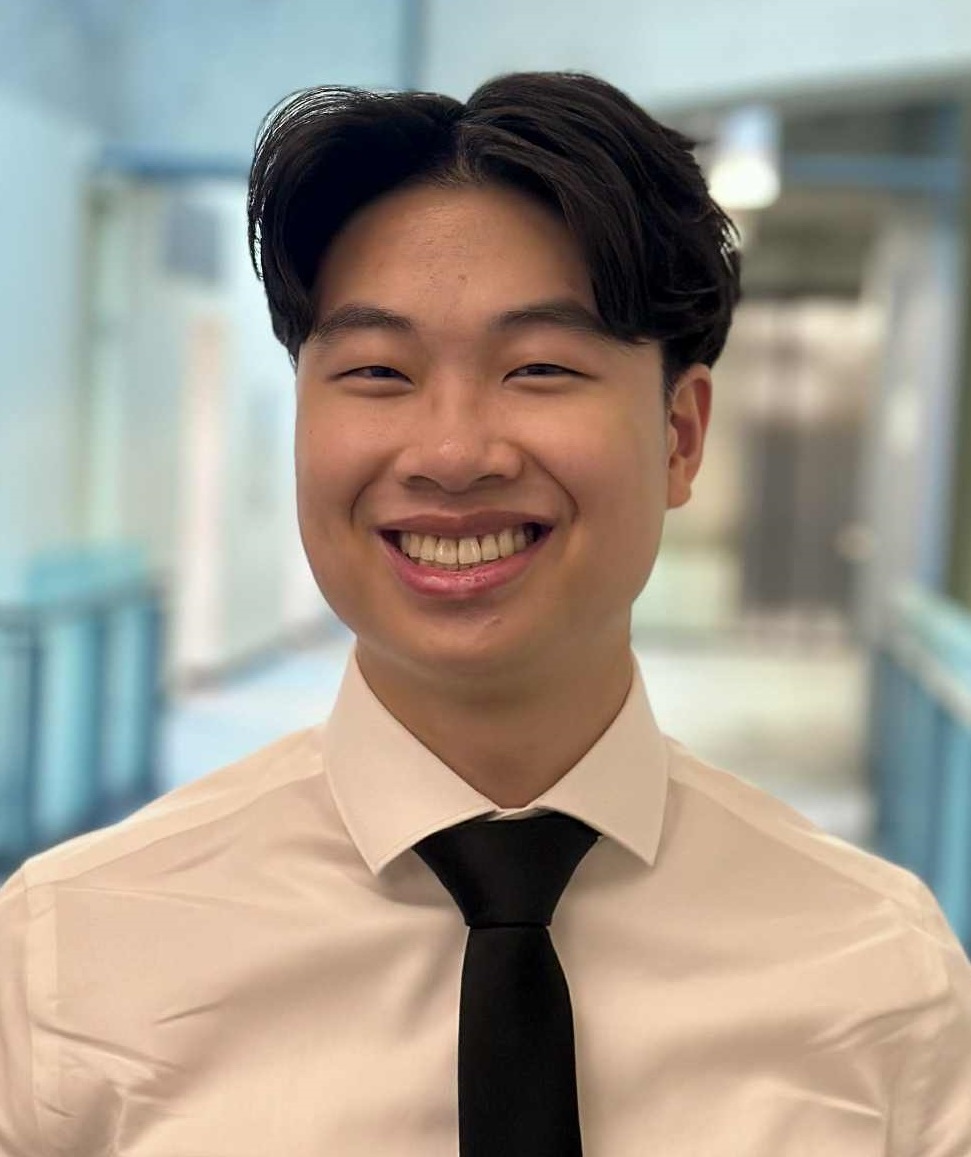}
\end{minipage}\hfill
\begin{minipage}[t]{0.85\textwidth}
\vspace{0pt}
\textbf{Alex Zhang} did his undergraduate degree in mechanical engineering at the University of Toronto, where he is currently pursuing his master's degree in robotics. His interests lie in autonomous control of robots and state estimation, particularly for mobile robots.
\end{minipage}

\vspace{10 pt}

\noindent
\begin{minipage}[t]{0.12\textwidth}
\vspace{0pt}
\includegraphics[width=0.8in,height=1in]{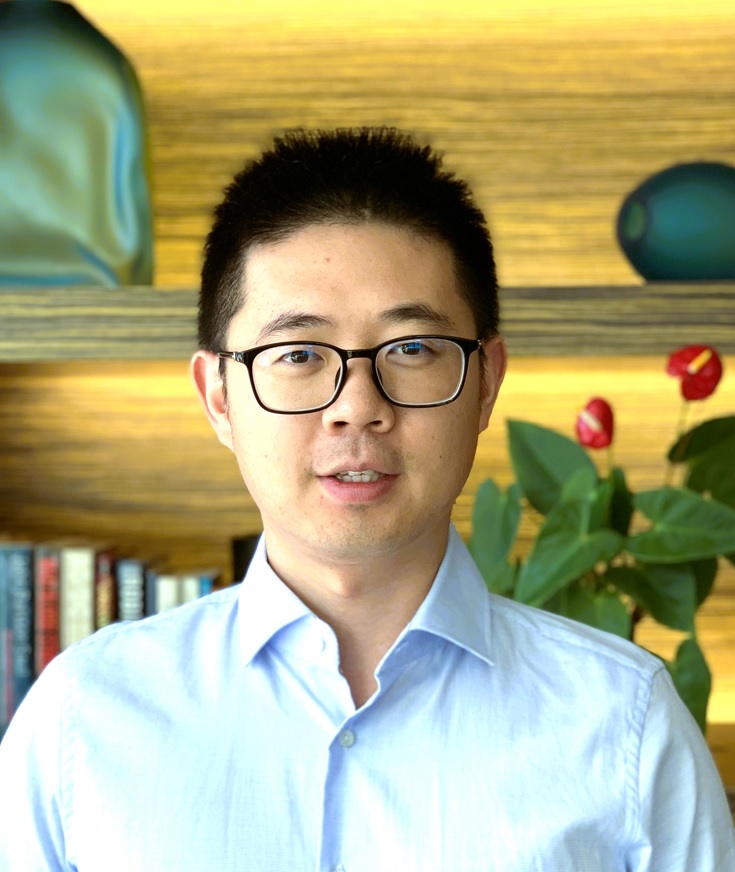}
\end{minipage}\hfill
\begin{minipage}[t]{0.85\textwidth}
\vspace{0pt}
\textbf{Changyan He} is an Assistant Professor in the School of Engineering at the University of Newcastle, Australia. His research focuses on the development of advanced robotic systems for minimally invasive surgical procedures, including magnetic microrobots, continuum manipulators, flexible sensors, and intelligent robotic control frameworks. He worked as a postdoctoral researcher at the University of Toronto and The Hospital for Sick Children (SickKids) from 2021 to 2023. He received his B.E. in Mechanical and Automation Engineering from Beijing Jiaotong University in 2015 and his Ph.D. from Beihang University in 2021. He was a visiting Ph.D. student at Johns Hopkins University from 2017 to 2019, where he conducted research in surgical robotics.
\end{minipage}

\vspace{10 pt}

\noindent
\begin{minipage}[t]{0.12\textwidth}
\vspace{0pt}
\includegraphics[width=0.8in,height=1in]{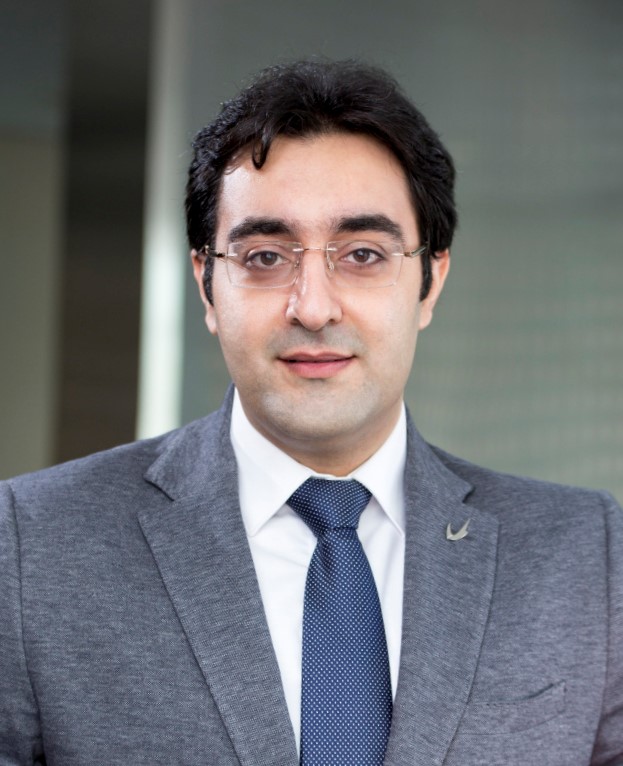}
\end{minipage}\hfill
\begin{minipage}[t]{0.85\textwidth}
\vspace{0pt}
\textbf{Amir Hooshiar} is an Assistant Professor of Surgery at the Department of Surgery since July 2023. He has also been the Founding Director of the Surgical Performance Enhancement and Robotics (SuPER) Centre. He has MSc (2009) and BSc (2007) in biomedical engineering from Tehran Polytechnique and PhD (2021) in mechanical engineering from Concordia University, Montreal. He received his postdoctoral training in surgical robotics at McGill (2021-2023). Dr. Hooshiar's research is focused on the design, development, and validation of autonomous and soft surgical robots as well as the development and validation of AI-based decision support systems for intraoperative guidance and surgical skill assessment.
\end{minipage}

\vspace{10 pt}

\noindent
\begin{minipage}[t]{0.12\textwidth}
\vspace{0pt}
\includegraphics[width=0.8in,height=1in]{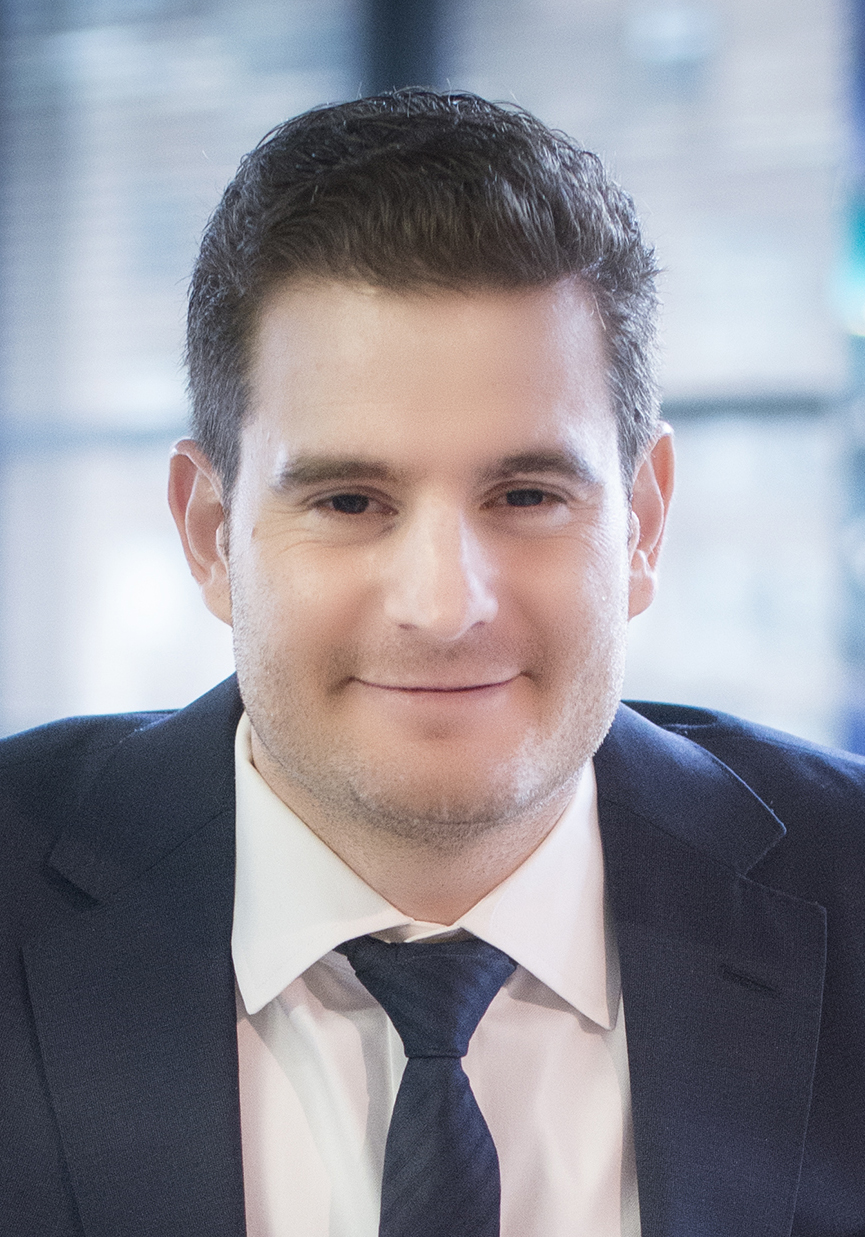}
\end{minipage}\hfill
\begin{minipage}[t]{0.85\textwidth}
\vspace{0pt}
\textbf{Dale J. Podolsky} is a craniofacial surgeon-scientist and mechanical engineer with dual degrees in physics and mechanical engineering. His academic focus centers on surgical robotics, medical device design, and translational innovation. Holding a doctorate in biomedical engineering, he develops robotic systems for minimally invasive surgical procedures. His research interests include device prototyping, haptics and simulation for surgical training, and image-guided navigation technologies. Through multidisciplinary collaboration, he integrates engineering and clinical expertise to create novel tools that enhance precision in craniofacial surgery.
\end{minipage}

\vspace{10 pt}

\noindent
\begin{minipage}[t]{0.12\textwidth}
\vspace{0pt}
\includegraphics[width=0.9in,height=1in]{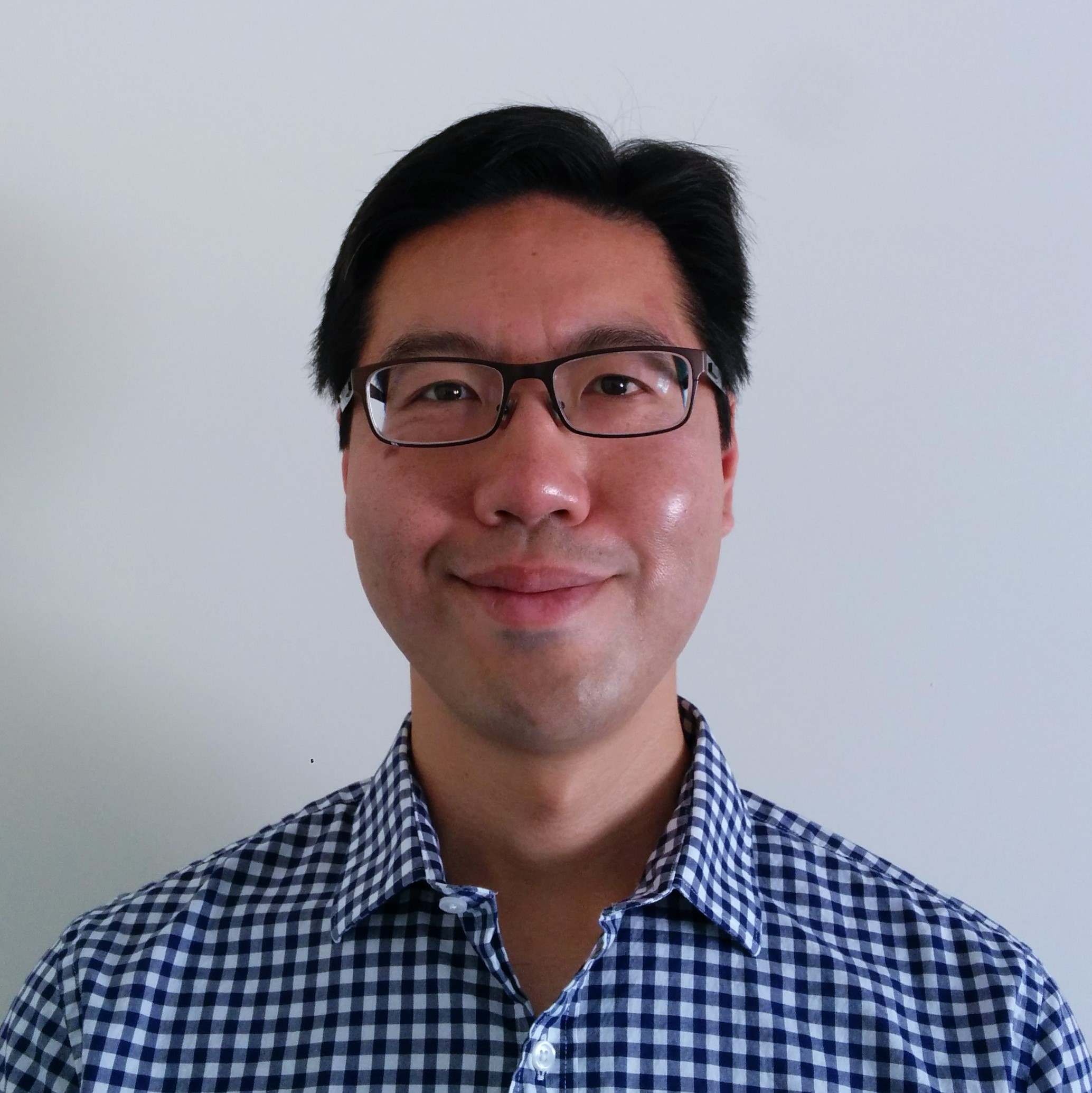}
\end{minipage}\hfill
\begin{minipage}[t]{0.85\textwidth}
\vspace{0pt}
\textbf{Thomas Looi} received his BASc degree in Engineering Science - Aerospace Option from the University of Toronto, MASc degree in Space Systems Engineering from the University of Toronto Institute of Aerospace Studies (UTIAS), and PhD from the Institute for Biomedical Engineering at the University of Toronto. He is the Posluns Innovator and Co-Lead for surgical robotics, 3D printing, and simulation at the Wilfred and Joyce Posluns Centre for Image Guided Innovation and Therapeutic Intervention in the Hospital for Sick Children. He is also an Assistant Professor (status only) in the Department of Mechanical and Industrial Engineering with a cross appointment in the Department of Otolaryngology, Head and Neck Surgery at the University of Toronto. His research interests are in medical and surgical robotics, 3D printing, microfabrication, surgical tool development, and minimally invasive surgery. 
\end{minipage}

\vspace{10pt}

\noindent
\begin{minipage}[t]{0.12\textwidth}
\vspace{0pt}
\includegraphics[width=0.9in,height=1in]{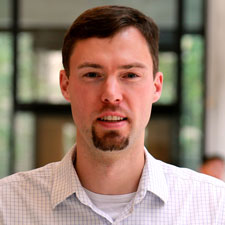}
\end{minipage}\hfill
\begin{minipage}[t]{0.85\textwidth}
\vspace{0pt}
\textbf{Eric Diller} received his B.S. and M.S. degrees in mechanical engineering from Case Western Reserve University in 2010 and his Ph.D. in mechanical engineering from Carnegie Mellon University in 2013. He is a Professor in the Department of Mechanical and Industrial Engineering at the University of Toronto, with cross-appointments in the Institute for Biomedical Engineering and the Robotics Institute, where he directs the Microrobotics Laboratory. His research focuses on small-scale robotics, particularly the fabrication and control of micro-scale devices using magnetic fields, robotic manipulation at the microscale, and smart materials. His contributions have been recognized with several awards, including Best Associate Editor at the 2015 IEEE International Conference on Automation and Robotics, the IEEE Robotics and Automation Society Early Career Award (2020), the Ontario Early Researcher Award (2018), the University of Toronto Connaught New Researcher Award (2017), and the Canadian Society of Mechanical Engineering's I.W. Smith Award (2018) for research in medical microrobotics.
\end{minipage}

\end{document}